\documentclass[review]{elsarticle}

\usepackage{amssymb}
\usepackage{amsmath}
\usepackage[linesnumbered, ruled]{algorithm2e}
\usepackage{algorithmic}
\usepackage{subfigure}
\usepackage{array}
\usepackage{booktabs}
\usepackage{multirow}
\usepackage{subfigure}
\usepackage{lineno}
\usepackage{hyperref}
\usepackage{color}
\usepackage{caption}
\begin{document}

\begin{frontmatter}



\title{Relation-aware graph structure embedding with co-contrastive learning for drug-drug interaction prediction}


\author[mymainaddress]{Mengying Jiang}
\ead{myjiang@stu.xjtu.edu.cn}

\author[mymainaddress]{Guizhong Liu\corref{mycorrespondingauthor}}
\cortext[mycorrespondingauthor]{Corresponding author}
\ead{liugz@xjtu.edu.cn}

\author[mymainaddress]{Biao Zhao}
\ead{biaozhao@xjtu.edu.cn}

\author[mysecondaryaddress]{Yuanchao Su}
\ead{suych3@xust.edu.cn}

\author[mymainaddress]{Weiqiang Jin}
\ead{weiqiangjin@stu.xjtu.edu.cn}

\address[mymainaddress]{School of Electronic and Information Engineering, Xi'an Jiaotong University, Xi'an 710049, China}
\address[mysecondaryaddress]{College of Geomatics, Xi'an University of Science and Technology, Xi'an 710054, China}

\begin{abstract}
Relation-aware graph structure embedding is promising for predicting multi-relational drug-drug interactions (DDIs). 
Typically, most existing methods begin by constructing a multi-relational DDI graph and then learning relation-aware graph structure embeddings (RaGSEs) of drugs from the DDI graph.
Nevertheless, most existing approaches are usually limited in learning RaGSEs of new drugs, leading to serious over-fitting when the test DDIs involve such drugs.
To alleviate this issue, we propose a novel DDI prediction method based on relation-aware graph structure embedding with co-contrastive learning, RaGSECo.
The proposed RaGSECo constructs two heterogeneous drug graphs: a multi-relational DDI graph and a multi-attribute drug-drug similarity (DDS) graph. 
The two graphs are used respectively for learning and propagating the RaGSEs of drugs, aiming to ensure all drugs, including new ones, can possess effective RaGSEs.
Additionally, we present a novel co-contrastive learning module to learn drug-pairs (DPs) representations.
This mechanism learns DP representations from two distinct views (interaction and similarity views) and encourages these views to supervise each other collaboratively to obtain more discriminative DP representations.
We evaluate the effectiveness of our RaGSECo on three different tasks using two real datasets. The experimental results demonstrate that RaGSECo outperforms existing state-of-the-art prediction methods. 
\end{abstract}



\begin{highlights}
\item We innovatively propose the propagation of relation-aware graph structure embeddings from known drugs to new drugs with similar features.
\item It is the first attempt to learn drug-pair feature representations based on cross-view contrastive learning, and the selected views are the interaction and similarity views.
\item We design a novel positive sample selection strategy that considers the underlying correlations between drug pairs, further enhancing the effectiveness of cross-view contrastive learning.
\end{highlights}


\end{frontmatter}



\section{Introduction} \label{Introduction}
Human diseases are often the result of complex biological processes, and single-drug treatments are often insufficient to cure them entirely \cite{SCAN, transfer}. Consequently, combination drug therapy has become attractive as it can reduce drug resistance and improve efficacy~\cite{NovelDP2019, DING2021618}. 
Nevertheless, the simultaneous use of multiple drugs can alter their properties and lead to adverse drug interactions, which can cause harm to patients~\cite{DBLP1}. 
Therefore, identifying potential DDIs is crucial for safe coadministration~\cite{DBLP2}. 
Traditional methods for detecting DDIs, such as biological or pharmacological assays, are labor-intensive, time-consuming, and expensive~\cite{HE2022247, LIAN20221}.
In contrast, computation-based methods are generally less expensive and can achieve higher accuracy than traditional methods~\cite{PPTPP}. 
In recent years, many computational methods have been proposed for predicting DDI events~\cite{pami2022}. 

Many existing works regard the drugs with known interactions as known drugs and refer to those without known interactions as new drugs~\cite{DDIMDL}. 
Typically, the lack of interaction information on new drugs makes it difficult for models to predict corresponding interactions accurately \cite{DDIMDL}.  
Nevertheless, predicting interactions involving new drugs is significant for the efficient development of new drugs~\cite{TMFUF}.
Therefore, many DDI prediction approaches concern this task~\cite{SSI-DDI2}.
For example, Deng et al.~\cite{DDIMDL} proposed the multimodal deep learning framework for predicting DDI events (DDIDML). DDIDML employs multiple biochemical attributes (such as enzymes, targets, pathways, and chemical substructures) to compute multiple similarities to build a deep neural network (DNN) model for multi-relational DDI prediction.
Zhang et al.~\cite{SFLLN} proposed the sparse feature learning ensemble method with linear neighborhood regularization (SFLLN) that combines multiple drug features like DDIDML and known DDIs to predict novel DDI.
Lin et al.~\cite{MDF-SA-DDI} proposed a DDI prediction method that jointly utilizes the multi-source drug fusion, the multi-source feature fusion, and the transformer (MDF-SA-DDI). The MDF-SA-DDI first utilizes four different drug fusion networks to encode latent features of DPs, then adopts transformers to perform latent feature fusion for representation learning of DPs. 
The aforementioned methods employ multiple attributes to learn representative drug embeddings and have achieved leading performance when predicting interactions involving new drugs.
Nevertheless, to ensure that the training and test sets have the same data distribution, these methods do not use multi-relational interaction information between drugs, limiting the DDI prediction performance.

Recently, self-supervised learning, which aims to derive supervised signals from the data itself spontaneously, has emerged as a promising strategy for effective representation learning \cite{WANG2022383}.
Among the various techniques under the umbrella of self-supervised learning, contrastive learning has attracted substantial attention \cite{MEN2023198}. Contrastive learning first extracts positive and negative samples from data and then maximize the similarities between positive samples while minimizing the similarities between negative samples\cite{csgnn}. 
In this way, contrastive learning can learn discriminative representations even without abundant labels. 
Despite the broad application of contrastive learning in computer vision \cite{LIU2022193} and natural language processing \cite{zhao2023chatagri}, its potential in the context of DDI prediction tasks has been scarcely explored.
Implementing contrastive learning to facilitate DDI prediction is by no means trivial and requires careful consideration of the task-specific characteristics and nuances of contrastive learning.

Based on the above discussion, the primary motivation for our work lies in enabling all drugs to distill effective RaGSEs, thereby facilitating DP representation learning and improving DDI prediction.
We propose a relation-aware graph structure embedding with co-contrastive learning framework for DDI prediction, abbreviated as RaGSECo, an end-to-end learning model.
Implementing our RaGSECo approach includes two main steps: drug embedding and DP representation learning.
The relation-aware graph structure embedding learning and propagation (RaGSELP) method is proposed for drug embedding.
RaGSELP constructs a multi-relational DDI graph by using known DDIs (training set).
Then, RaGSELP learns RaGSEs of known drugs by aggregating their neighbor’s features under different relations.
Inspired by an assumption that similar drugs may interact with the same drugs~\cite{{SHANG202180},{MDF-SA-DDI},{DDIMDL}}, RaGSELP constructs a multi-attribute DDS graph.
Within this DDS graph, RaGSELP learns embeddings of new drugs by aggregating their neighbor’s RaGSEs, aiming to enable all drugs to possess effective RaGSEs.
Furthermore, we incorporate multi-view representation learning and cross-view contrastive Learning to present a novel co-contrastive learning mechanism to learn DP representations.
Unlike previous contrastive learning, which contrasts the original and the corrupted networks, we design two distinct views for DP representation learning: interaction and similarity views.
Specifically, we leverage the RaGSEs of drugs to generate the DP representations under the interaction view, and we employ similarities between drugs to generate DP representations under the similarity view.
The interaction view primarily utilizes known interaction relationships between drugs.
The similarity view can discover inherent connections between drugs and facilitate inferring the potential therapeutic effects, toxicity reactions, or drug interactions of new drugs. 
Therefore, the two views are complementary.
Moreover, we consider underlying correlations between DPs to design a novel positive selection strategy to enhance cross-view contrastive learning.
With the training, these two views supervise each other collaboratively and learn more discriminative DP representations.

The main contributions of our work can be summarized as follows:
\begin{itemize}
\item[]
$\bullet$ Propagating the RaGSEs of known drugs to new drugs with similar features is innovatively proposed on relation-aware-based methods. 
This significantly improves DDI prediction performances of relation-aware-based methods, especially in predicting DDIs involving new drugs.
\end{itemize}
\begin{itemize}
\item[] 
$\bullet$ To our best knowledge, this is the first attempt to perform the DP-level cross-view contrastive learning. 
More discriminative DP representations can be learned by co-contrastive learning based on a cross-view manner.
\end{itemize}
\begin{itemize}
\item[]
$\bullet$ The proposed RaGSECo ingeniously contrasts and incorporates two views of the drug information network (interaction and similarity views), enabling DP to capture both the known and the potential interactions.
\end{itemize}

\section{Related Work}
\label{related work}

We review previous studies relevant to this work in three areas: 
relation-aware graph structure embedding, contrastive learning, and drug feature extraction.

\subsection{Relation-Aware Graph Structure Embedding}\label{RaGNN}

The methods based on relation-aware graph structure embedding learning pay attention to the topology of the graph, typically learning entity embeddings by aggregating information from neighboring entities under different relations \cite{RANEDDI}. 
The practicability of these methods on multi-relational DDI predictions makes these methods attract considerable attention.
Most existing methods use graph neural networks (GNNs), including relational graph convolutional networks (RGCNs) \cite{rgcn} and graph attention networks (GATs) \cite{gat}.
For example, Hong et al.~\cite{LaGAT} used the GAT to propose the link-aware graph attention network (LaGAT) that learns drug embedding by aggregating features of neighbors from different attention pathways via different DDI event types, where the attention weights depend on embedding representations of drugs and their neighbors.
Wang et al.~\cite{GOGNN} proposed the graph of graphs neural network (GoGNN) that captures the information on drug structure and multi-relational drug interactions in a hierarchical way to learn drug embedding for DDI prediction.
Yu et al.~\cite{RANEDDI} proposed the relation-aware network embedding for DDI prediction (RANEDDI) that considers both the multi-relational information between drugs and the relation-aware network structure information of drugs to obtain the drug embedding for DDI prediction.
Yu et al.~\cite{STNN-DDI} proposed the substructure-aware tensor neural network model for DDI prediction (STNN-DDI) that learns a 3-D tensor of (substructure, substructure, interaction type) triplets, which characterizes a substructure–substructure interaction (SSI) space.
The aforementioned methods take advantage of interaction information and perform well in the multi-relational DDI prediction task. Nevertheless, the absence of interaction information on new drugs limits the performance of predicting DDIs involving new drugs.

\subsection{Contrastive Learning}\label{ContrastiveLearning} 

The approaches based on contrastive learning learn representations by contrasting positive pairs against negative pairs have achieved considerable success across various domains \cite{ss1}.
In this section, we mainly focus on reviewing the contrastive learning methods related to DDI prediction tasks.
Zhao et al.~\cite{csgnn} constructed original and corrupted networks, then minimized the mutual information between outputs of original and corrupted networks, and maximized the mutual information between outputs from only the original or corrupted network.
Wang et al.~\cite{BioERP} proposed a self-supervised meta-path detection mechanism to train a deep transformer encoder model that can capture the global structure and semantic features in heterogeneous biomedical networks.
Gao et al.~\cite{GSL} designed the drug-disease associations view and drug or disease similarity view, then maximized the mutual information between the two views.
Zhuang et al.~\cite{DMVDGI} learned high-level drug representations containing graph-level structural information by maximizing the local-global mutual information.
Although many works have been proposed to learn high-level drug embeddings by contrastive learning, little effort has been made toward DP-level contrastive learning. Nevertheless, learning high-level DP representations is more practical for the DDI prediction task.

\subsection{Drug Feature Extraction}\label{dfe}

Drug feature extraction is crucial for model training \cite{MDF-SA-DDI}. 
Zhu et al.~\cite{pami2022} took into account eight drug attributes (molecular
substructure, target, enzyme, pathway, side effect, phenotype, gene, and disease) to extract features.
Deng et al. \cite{DDIMDL} jointly consider targets, enzymes, and chemical substructures, which achieved outstanding performance~\cite{MDF-SA-DDI}. 
Typically, each attribute is associated with a set of descriptors. A drug can be denoted as a binary feature vector where the element (1 or 0) indicates the presence or absence of the corresponding descriptor~\cite{MDF-SA-DDI}.
These sparse binary feature vectors typically have high dimensionalities.
In general, high dimensional inputs can be resource-intensive and induce the curse of dimensionality, leading to extremely poor performance in some cases~\cite{MDF-SA-DDI}.
Therefore, given that similar drugs may interact with the same drugs, Deng et al. \cite{DDIMDL} opted to use the Jaccard similarities calculated from the binary feature vectors to define drugs rather than the binary feature vector itself. The Jaccard similarity is calculated as follows:
\begin{equation}
\operatorname{Jaccard}(U, V)=\frac{|U \cap V|}{|U \cup V|}
\label{jaccard}
\end{equation}
where $U$ and $V$ are the descriptors of two drugs under a specific attribute. Herein, $|U \cap V|$ is the cardinality of the intersection between $U$ and $V$, and $|U \cup V|$ is the union.
\section{Methodology}

\subsection{RaGSECo framework}\label{framework}

\begin{figure}[t]
\centering
\includegraphics[width=3.5in,height=1.5in]{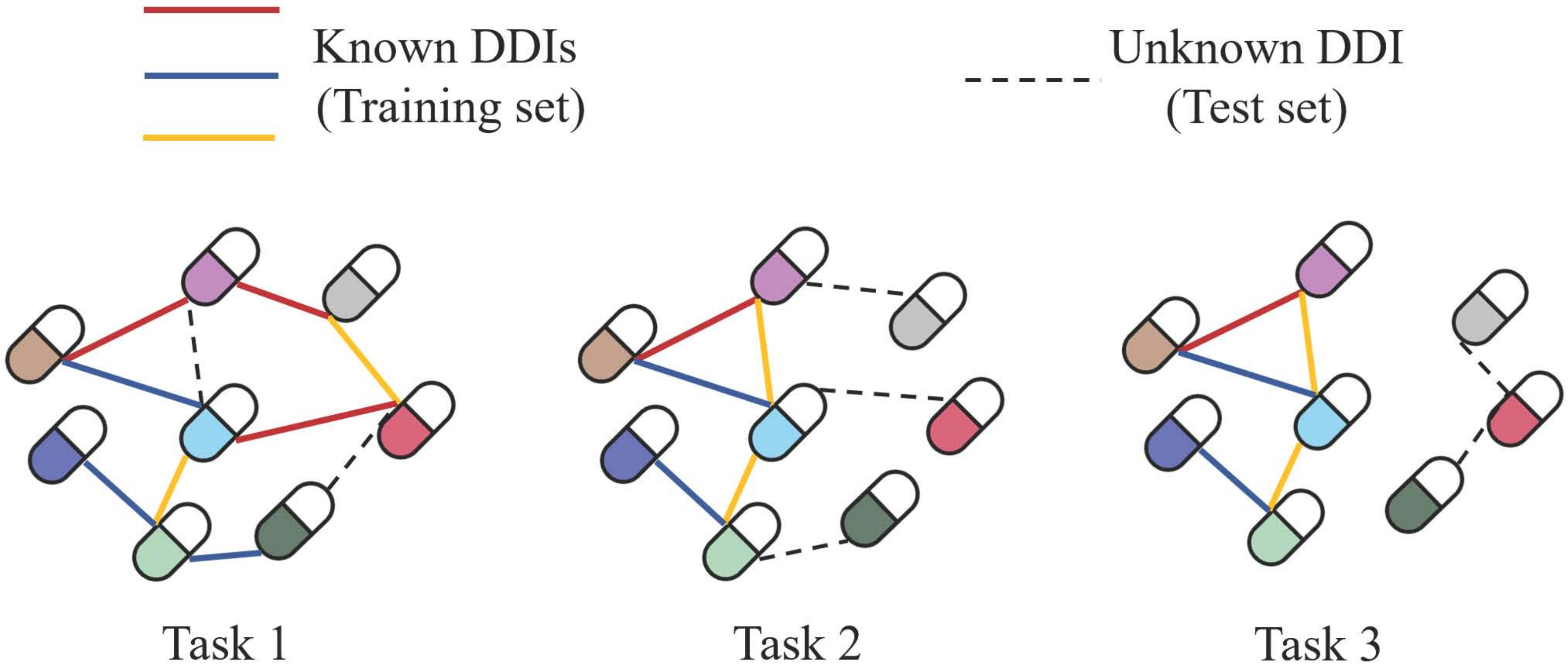}
\caption{Examples of construction strategies for three different test sets. The nodes represent drugs. Solid edges indicate known DDI interactions (training set), while dotted edges represent the prediction task (test set). The edges with different colors signify various interaction types.}
\label{3task}
\end{figure}

The RaGSECo is proposed for multi-relational DDI prediction, which can be regarded as a multi-class classification task. 
Based on three different experimental settings, the multi-relational DDI prediction task can be further partitioned into three different tasks, defined as follows:
\begin{itemize}
\item[]
$\bullet$ Task 1: predicting unobserved interaction events between known drugs.
\end{itemize}
\begin{itemize}
\item[]
$\bullet$ Task 2: predicting interaction events between known drugs and new drugs.
\end{itemize}
\begin{itemize}
\item[]
$\bullet$ Task 3: predicting interaction events between new drugs.
\end{itemize}

These three tasks are vividly illustrated in Fig.~\ref{3task}.
As the figure illustrates, the training and test sets contain the same drugs in Task 1.
In Task 2, half of the drugs involved in the test set appear in the training set. 
In Task 3, the training and test sets possess different drugs.
Therefore, from Task 1 to Task 3, the prediction difficulties increase in turn.

\begin{figure*}[t]
\centering
\includegraphics[width=4.7in,height=2.15in]{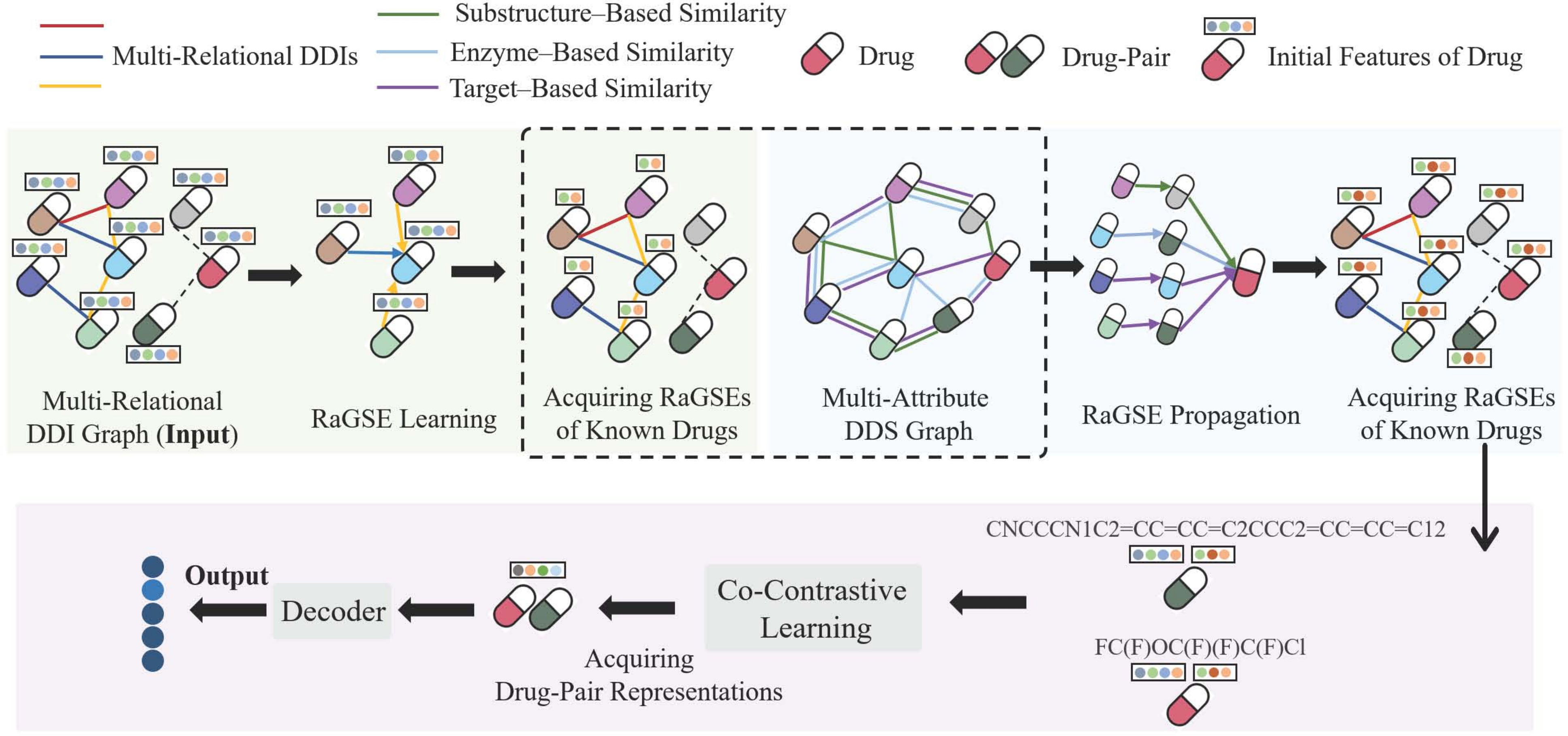}
\caption{Flowchart of the proposed RaGSECo, involving two main parts: RaGSELP ($\bf{Upper}$) and co-contrastive learning ($\bf{Bottom}$).
In the RaGSELP method, we first learn the RaGSEs of known drugs by propagating their neighbor’s features under different relations within the DDI graph. 
Subsequently, we propagate the learned RaGSEs to the connected new drugs within the DDS graph to ensure that all drugs possess effective RaGSEs. 
Afterward, the co-contrastive learning module takes in multiple features (initial features, RaGSEs, and SMILES strings) of two drugs and then generates the feature representations of the DP.
Finally, the Decoder calculates the classification probability distributions of the DP.
}
\label{flow} 
\end{figure*}

Taking Task 3 as an example, the framework of our RaGSECo is illustrated in Fig.~\ref{flow}.
The RaGSECo constructs two heterogeneous drug graphs: a multi-relational DDI graph and a multi-attribute DDS graph.
In the DDS graph, nodes denote drugs, and edges represent similarities between drugs.
The different edge colors indicate similarities under varying attributes (targets, enzymes, and chemical substructures).
Consequently, in the DDS graph, each pair of drugs can have up to three types of edges.
In contrast, in the DDI graph, each pair of drugs has only up to one type of edge.
The framework of our RaGSECo contains two primary parts: drug embedding and DP representation learning.
The RaGSELP method is proposed to learn drug embedding, and the cross-view module is developed for learning DP representation for DDI prediction.

\subsection{RaGSELP}\label{RaGSELP}

The RaGSELP method is proposed to learn drug embedding.
RaGSELP contains two parts: RaGSE learning (RaGSEL) and RaGSE propagation (RaGSEP).
RaGSEL learns the RaGSEs of known drugs, and RaGSEP propagates the RaGSEs of known drugs to new drugs with similar features.

\subsubsection{RaGSEL}\label{RaGSEL}
We construct a multi-relational DDI graph $\mathcal{G}=(\mathcal{D}, \mathcal{E}, \mathcal{R})$, where nodes represent drugs, and edges denote labeled interactions.
Herein, $\mathcal{D}$ is the set of all drugs (including known and new drugs), 
and $|\mathcal{D}|=N$ represents the number of drugs.
Let ${\bf X}= \left\{{\bf x}_i\right\}_{i=1}^N \in \mathbb{R}^{N \times d}$ be the initial feature matrix of drugs,
derived from Section~\ref{dfe}.
$\mathcal{R}$ is the set of interaction event types, and $|\mathcal{R}|=R$ is the number of interaction type.
$\mathcal{E}=\left\{\mathcal{E}_r\right\}_{r=1}^R$ is the set of known interactions, and $\mathcal{E}_r$ represents the set of interactions under interaction type $r$. 
Let $\mathcal{A}=\left\{{\bf A}_r\right\}_{r=1}^R\in \mathbb{R}^{N \times N \times R}$ be the multi-relational adjacency tensor, where ${\bf A}_r\in \mathbb{R}^{N \times N}$ represents the adjacency matrix under interaction relation $r$.
Let $A_{r(i,j)}$ $(i,j = 1,\dots,N)$ be the element of $\bf{A}_{r}$,
$A_{r(i,j)}=1$ if $(i,j) \in \mathcal{E}_{r}$ and $A_{r(i,j)}=0$ if $(i,j) \notin \mathcal{E}_{r}$.

Subsequently, we use the R-GCN layer \cite{rgcn} to learn the RaGSEs of drugs from the multi-relational DDI graph.  
The forward propagation function is defined as follows:
\begin{equation}
\mathbf{h}_i=\sigma\left(\sum_{r \in \mathcal{R}} \sum_{j \in \mathcal{N}_i^r} \frac{\hat{A}_{r(i,j)}}{R_i} \mathbf{x}_j\mathbf{W}_r + \mathbf{x}_i\mathbf{W}_o\right)\\
\label{sub}
\end{equation}
where $\mathbf{h}_i\in \mathbb{R}^{1 \times d'}$ represents the RaGSEs of drug $i$,
$\mathbf{x}_i \in \mathbb{R}^{1 \times d}$ are the initial feature vector of drug $i$, $i \in \mathcal{D}$. 
Herein, $\mathbf{W}_r\in \mathbb{R}^{d \times d'}$ represents the relation-specific weight matrix and the adoption of a set of $\mathbf{W}_r$ $(r = 1,\dots, R)$ supports multiple edge types.
On the contrary, $\mathbf{W}_0\in \mathbb{R}^{d \times d'}$ is a single weight matrix regardless of relations.
Here, $\mathcal{N}_i^r$ represents the set of drugs connected to drug $i$ under relation $r$.
$\sigma(\cdot)$ is an element-wise activation function: $\operatorname{ReLU}(\cdot)=\max (0, \cdot)$.
To normalize the incoming messages of each drug, we use $\left\{R_{i}\right\}_{i=1}^N$ to define a set of normalized constants. Here, $R_i$ equals the number of interaction types in which drug $i$ is involved.
$\hat{A}_{r(i,j)}$ is the aggregation weight assigned to the neighboring drug $j$.
Specifically, $\hat{A}_{r(i,j)}$ denotes the element in $\hat{\bf A}_{r}$. 
The calculation of $\hat{\bf A}_{r}$ is based on a classic graph-based normalization method~\cite{GCN}. 
\begin{equation}
\hat{\bf A}_{r} ={\bf D}_r^{-1/2}{\bf A}_{r}{\bf D}_r^{-1/2}
\label{norm}
\end{equation}
where ${\bf D}_r$ is the degree matrix of ${\bf A}_r$ and a diagonal matrix. 

Herein, $\mathbf{h}_i$ can represent relation-aware graph structure information of drug $i$ if drug $i$ is known. 
Nevertheless, if drug $i$ is new, $\mathbf{h}_i$ would be influenced only by drug $i$ itself. 
Consequently, directly using the output of Eq. (\ref{sub}) as the drug embeddings for DDI prediction could potentially result in severe over-fitting issues for Tasks 2 and 3.

\subsubsection{RaGSEP}\label{RaGSEP}

Based on the assumption that similar drugs may interact with the same drugs~\cite{MDF-SA-DDI, DDIMDL}, we propose a strategy that propagates the RaGSEs of known drugs to new drugs with similar features. This strategy can effectively overcome the over-fitting issues that arise in Tasks 2 and 3.
To facilitate the propagation of embeddings, 
we construct a multi-attribute DDS graph $\mathcal{G}_{s}=(\mathcal{D}, \mathcal{E}, \mathcal{S})$, where nodes represent drugs, and edges denote Jaccard similarities between drugs under various attributes.
Herein, we still employ these three biological attributes (i.e., chemical substructure, enzyme, and target).
In the DDS graph, $\mathcal{D}$ is the set of nodes,
with ${\bf H}=\left\{{\bf h}_{i}\right\}_{i=1}^N$ as the feature matrix of nodes, which is the output of Eq.~(\ref{sub}).
$\mathcal{E}$ represents the set of edges.
$\mathcal{S}$ is the set of similarity types, and $|\mathcal{S}|=3$ is the number of similarity types. 
Let $\mathcal{A}=\left\{{\bf A}_s, {\bf A}_e, {\bf A}_t\right\}\in \mathbb{R}^{N \times N \times 3}$ be the adjacency tensor, where ${\bf A}_s$, ${\bf A}_e$, and ${\bf A}_t$ represent the substructure-based, enzyme-based, and target-based adjacency matrix, respectively.   

To normalize the incoming messages, these three adjacency matrices also need to be normalized. 
The normalization method is referred to as Eq. (\ref{norm}).
Let $\hat{\bf A}_{s}$, $\hat{\bf A}_{e}$, and $\hat{\bf A}_{t}$ be the normalized adjacency matrices.
We use these normalized adjacency matrices to propagate the learned RaGSEs to the strongly associated drugs within the DDS graph. 
The propagation procedure can be expressed as follows:
\begin{equation}
\label{propa1} 
{\bf H}_{s} = \sigma\left(\hat{\bf A}_{s}^{n} {\bf H} {\bf W}_s \right)
\end{equation}
\begin{equation}
\label{propa2}  
{\bf H}_{e} = \sigma\left(\hat{\bf A}_{e}^{n}{\bf H} {\bf W}_e \right) 
\end{equation}
\begin{equation}
\label{propa3}  
{\bf H}_{t} = \sigma\left(\hat{\bf A}_{t}^{n} {\bf H} {\bf W}_t \right)
\end{equation}
\begin{equation}
\label{propa4} 
{\bf H}_{embed} = {\bf H}_{s}+{\bf H}_{e}+{\bf H}_{t} 
\end{equation}
where $\hat{\bf A}^{n}_{s} \bf{H}$, $\hat{\bf A}^{n}_{e} \bf{H}$, and $\hat{\bf A}^{n}_{t} \bf{H}$ represent graph convolutions on the DDS graph.
$n$ is a hyperparameter, and $\hat{\bf A}_{s}^{n}$ is $\hat{\bf A}_{s}$ to the power of $n$.
The same definition is extended to $\hat{\bf A}_{e}^{n}$ and $\hat{\bf A}_{t}^{n}$.
This strategy allows RaGSECo to capture higher-order neighbor information flexibly.
$\mathbf{W}_s$, $\mathbf{W}_e$, $\mathbf{W}_t \in \mathbb{R}^{d' \times d'}$ are the attribute-specific trainable weight matrices.
$\mathbf{H}_{s}$, $\mathbf{H}_{e}$, and $\mathbf{H}_{t}$ are the propagation results along with different pathways. 
$\mathbf{H}_{embed}=\left\{{\bf h}_{embed, i}\right\}_{i=1}^N\in \mathbb{R}^{N \times d'}$ is the sum of the three propagation results and represents the final embeddings of drugs.
Particularly, ${\bf h}_{embed, i}$ can represent effective RaGSEs of drug $i$, whether drug $i$ is known or new.

\begin{figure*}[t]
\centering
\includegraphics[width=4.7in,height=1.5in]{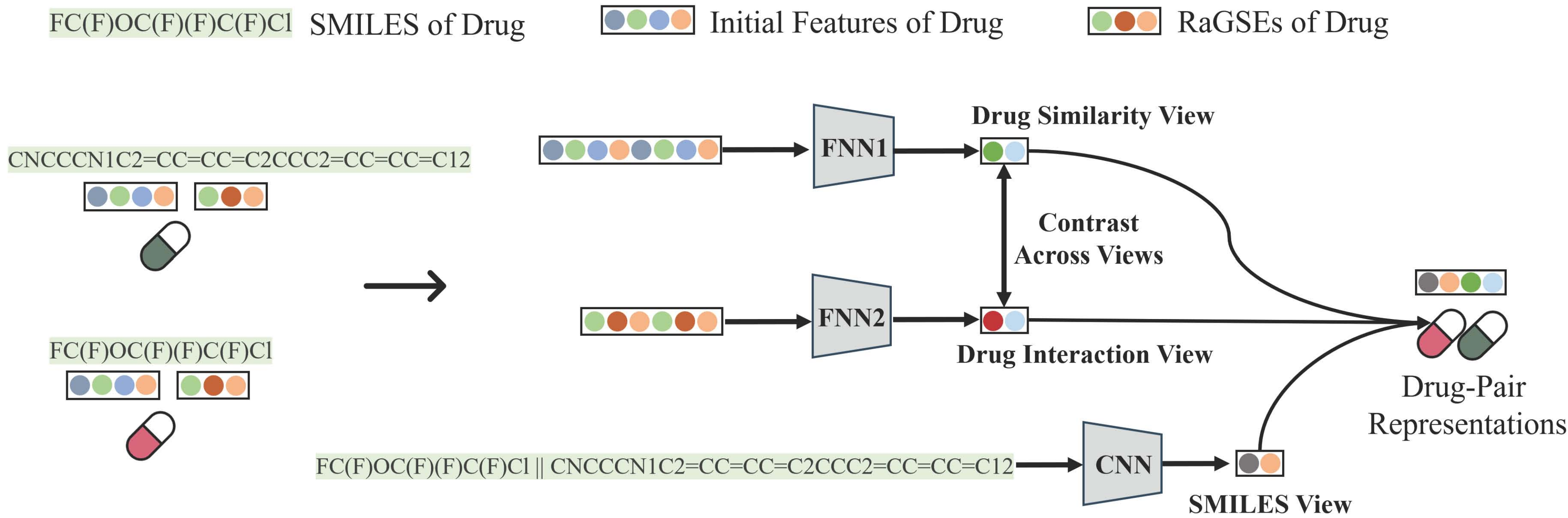}
\caption{Illustrations of the co-contrastive learning module. 
This module accepts the initial feature vectors, RaGSEs, and SMILES string of two drugs as the input and generates the DP representations. 
The initial feature vectors of the two drugs are concatenated and fed into an FNN, resulting in the DP representations under the similarity view.
In a similar fashion, another FNN is utilized to generate the DP representations from the interaction view. 
Meanwhile, a Convolutional Neural Network (CNN) is employed to obtain the SMILES-based DP representations. 
Additionally, a cross-view contrastive mechanism facilitates collaborative supervision between similarity and interaction views, thereby promoting the learning of more discriminative DP representations.
}
\label{Drug-Pair}
\end{figure*}

\subsection{Co-Contrastive Learning}\label{DPRL}

The co-contrastive learning module, as illustrated in Fig.~\ref{Drug-Pair}, is designed to learn DP representations.
This module consists of two primary components: multi-view representation learning and collaborative contrastive optimization.
The first component learns DP representations from multiple views.
The second component facilitates a cooperative process in which these views optimize and supervise each other.

\subsubsection{Multi-View Representation Learning}\label{initial features}

To increase data diversity, facilitate gradient propagation, and generate discriminative DP representations,
we employ multiple drug features containing RaGSEs, initial features, and Simplified-Molecular-Input-Line-Entry-System (SMILES) string.

SMILES string is a line notation that uses a predefined set of rules to describe the structure of compounds, and each drug has its own SMILES string \cite{IIFDTI}.
The characters of a SMILES string represent chemical atoms or chemical bonds, and the lengths of the SMILES strings are unconstrained \cite{SSI–DDI}.
In this work, we convert each SMILES string to a $p\times q$ dimensional feature matrix for simplicity, where $p$ denotes the number of characters and $q$ is the unified length of the SMILES string \cite{DeepPurpose}.
Accordingly, the columns of the SMILES-based feature matrix are one-hot vectors.
Since the initial lengths of the SMILES strings are flexible,
we will cut off the extra part if the actual length of the SMILES string is greater than $q$, and we will pad zeros if the actual length is less than $q$.
The $p$ and $q$ are set as $64$ and $100$ according to Huang et al.~\cite{DeepPurpose}.

Let $Q$ be the number of the known DDIs.
Considering computational burdens, we adopt a batch-wise scheme to train the RaGSECo.
Let $K$ be the number of DPs in a batch.
Given DP $k$ $(k=1,\dots\, K)$ that involves two drugs $i$ and $j$ $(i,j=1,\dots\, N;i \neq j)$.
Let ${\bf{S}}_i$, ${\bf{S}}_j\in \mathbb{R}^{p \times q}$ be the SMILES-based feature matrices of drugs $i$ and $j$.
${\bf{h}}_{embed,i}$ and ${\bf{h}}_{embed,j}$ are the RaGSEs of two drugs.
${\bf x}_i$ and ${\bf x}_j$ are initial feature vectors.
Subsequently, we employ two feedforward neural networks (FNNs) to process the RaGSEs and initial features of DPs, respectively.
In addition, a CNN encodes SMILES strings of DPs.
The employed CNN is a multi-layer 1-D CNN followed by a global max pooling layer, and with reference to \cite{DeepPurpose}.
Accordingly, three features of DP $k$ are encoded as follows:
\begin{equation}
\label{views1} 
{\bf p}_{k}^{initi}=\operatorname{FNN1}({\bf x}_i||{\bf x}_j,\Theta_{\mathrm{FNN1}})
\end{equation}
\begin{equation}
\label{views2} 
{\bf p}_{k}^{embed}=\operatorname{FNN2}({\bf h}_{embed,i}||{\bf h}_{embed,j},\Theta_{\mathrm{FNN2}})
\end{equation}
\begin{equation}
\label{views3} 
{\bf p}_{k}^{smile}=\operatorname{CNN}(\mathbf{S}_i||\mathbf{S}_j,\Theta_{\mathrm{CNN}}).
\end{equation}
where the symbols $\Theta_{\mathrm{FNN1}}$, $\Theta_{\mathrm{FNN2}}$, and $\Theta_{\mathrm{CNN}}$ represent the trainable weights involved in FNN1, FNN2, and CNN, respectively. 

The initial feature vectors of drugs are denoted by Jaccard similarity scores between drugs.
Thus, ${\bf p}_{k}^{initi} \in \mathbb{R}^{1 \times d^{FNN}}$ can be considered as the DP feature representations under the similarity view.
Meanwhile, the drug RaGSEs primarily concentrate on interaction information between drugs.
As such, ${\bf p}_{k}^{embed} \in \mathbb{R}^{1 \times d^{FNN}}$ can be perceived as the DP representations under the interaction view.

\subsubsection{Collaboratively Contrastive Optimization}\label{Collaboratively}

The interaction view mainly focuses on known interaction relationships between nodes, while the similarity view can infer potential therapeutic effects, adverse reactions, or drug interactions of new drugs by discovering underlying drug correlations.
Therefore, these two views are complementary and mutually reinforcing.
Herein, we define positive and negative samples for contrastive learning.
In this paper, given the feature representations of a DP under the interaction view, we can simply define its feature representations under the similarity view as the positive sample.
Nevertheless, our analysis of the employed datasets suggests that there may be underlying correlations between DPs. Consequently, we propose a novel positive selection strategy.

This paper utilizes two real-world, multi-relational DDI datasets: Dataset 1 and Dataset 2. 
In Dataset 1, the known DDIs are categorized into 65 types of DDI events, 
while the number of interaction event types in which each drug is involved ranges from 1 to 20.
The median of the event type counts is 10, significantly lower than 65. 
Similar findings are also observed in Dataset 2.
These findings suggest that a particular drug is likely to associate with specific interaction event types strongly.
Thus, the interaction event types a drug participates in can be considered significant characteristics of the drug.
We refer to these significant characteristics as interaction characteristics of drugs.
Let ${{\bf t}_i}\in {\mathbb R}^{1 \times R}$ represent the interaction characteristics of drug $i$, where $R$ is the number of interaction event types.
The vector ${\bf t}_i$ is binary, and its nonzero elements represent the interaction event types the drug $i$ participates in.
Given a DP $k$ that involves two drugs $i$ and $j$, the interaction characteristics of DP $k$ are defined as ${\bf c}_k={\bf t}_i + {\bf t}_j$.
Based on this, we propose a new positive selection strategy: if two DPs exhibit similar interaction characteristics, they are designated as positive samples.
One advantage of such a strategy is that the selected positive samples may reflect the underlying interaction tendency of the target DP.
For DPs $k$ and $q$, we calculate the cosine similarity of their interaction characteristics as follows:
\begin{equation}
\label{cosine} 
{S}_{k,q} = \frac{{\bf c}_k \cdot {\bf c}_q}{\|{\bf c}_k \| \|{\bf c}_q\|}.
\end{equation}

Given the threshold values $t_{pos}$ and $t_{neg}$, where $t_{pos}>t_{neg}$.
DP $q$ is deemed a positive sample of DP $k$ if ${ S}_{k,q} \geq t_{pos}$, and conversely, is considered a negative sample if ${ S}_{k,q} \leq t_{neg}$.
Let $\mathcal{P}$ represent the set of positive pairs of one batch, 
and $\mathcal{N}$ be the set of negative pairs.
We then model a self-supervised learning task using the standard binary cross-entropy loss.
With this, we present the co-contrastive learning loss function as follows:
\begin{equation}
\label{coss1}
\begin{aligned}
\ell_{ss1}= & -\frac{1}{|\mathcal{P}|+|\mathcal{N}|}\left(\sum_{(k,q)\in\mathcal{P}} \log \Psi\left({\bf p}_{k}^{embed}, {\bf p}_{q}^{initi}\right) \right. \\
& \left.+\sum_{(k,q)\in\mathcal{N}} \log \left(1-\Psi\left({\bf p}_{k}^{embed}, {\bf p}_{q}^{initi}\right)\right) \right)
\end{aligned}
\end{equation}
where the contrastive discriminator $\Psi(\cdot,\cdot)$ is instantiated as $\sigma({\bf p}_{k}^{embed} {\bf W} {{\bf p}_{q}^{initi}}^{T})$.
${\bf W}$ is a learnable weight matrix.
The activation function $\sigma$ is the sigmoid function, which produces a score representing the probability of being a positive sample.
The co-contrastive learning mechanism maximizes the mutual information of positive pairs under the interaction and similarity views.
To stabilize the co-contrastive learning process and facilitate two views supervising each other, we present another co-contrastive learning loss function:
\begin{equation}
\label{coss2}
\begin{aligned}
\ell_{ss2}= & -\frac{1}{|\mathcal{P}|+|\mathcal{N}|}\left(\sum_{(k,q)\in\mathcal{P}} \log \Psi\left({\bf p}_{k}^{initi}, {\bf p}_{q}^{embed}\right) \right. \\
& \left.+\sum_{(k,q)\in\mathcal{N}} \log \left(1-\Psi\left({\bf p}_{k}^{initi}, {\bf p}_{q}^{embed}\right)\right) \right).
\end{aligned}
\end{equation}

In conclusion, to fully utilize all significant information and learn more discriminative DP representations, we define the final DP representations ${\bf p}_{k}$ as follows:
\begin{equation}  
\label{dpfinal1} 
{\bf p}_{k}^{add} = {\bf p}_{k}^{initi}+{\bf p}_{k}^{embed}+{\bf p}_{k}^{smile} 
\end{equation}
\begin{equation}  
\label{dpfinal2}
{\bf p}_{k} = ||\left({\bf p}_{k}^{initi},{\bf p}_{k}^{embed}, {\bf p}_{k}^{smile},{\bf p}_{k}^{add}\right).
\end{equation}
where the symbol $||$ denotes the concatenation.

\subsection{Loss Function}\label{loss}
We construct a Decoder by two fully connected layers that map ${\bf p}_k$ into the probability distribution space ${\bf y}_k\in \mathbb{R}^{1 \times R}$:
\begin{equation}
{\bf y}_k=\operatorname{Decoder}\left({\bf p}_k;\Theta_{\mathrm{Decoder}}\right)
\label{predict}
\end{equation}
where $\Theta_{\mathrm{Decoder}}$ is the trainable parameters. 
The first fully connected layer is followed by an activation function GeLu \cite{gelu}, a batch normalization layer \cite{Batchnorma}, and a dropout layer.
The second fully connected layer is followed by a softmax function.

DDI events prediction is a multi-classification task. Thus, we choose the cross-entropy loss function as our classification loss function:
\begin{equation}
\mathcal{L}_{ce}= -\sum_{k=1}^{K} \sum_{r=1}^{R} \hat{y}_{k,r} \log \left(y_{k,r}\right)
\label{lossce}
\end{equation}
where $\hat{y}_{k,r}$ represents the $r$-th element of $\hat{\bf{y}}_{k}$, which corresponds to the ground-truth vector (one-hot encoding) of the DP $k$. $K$ denotes the number of DDI samples in a single batch for training.
On the other hand, $y_{k,r}$ denotes the predicted probability score of the DP $k$ under the class $r$, corresponding to the $r$-th element of ${\bf y}_k$.

To emphasize the importance of the classification loss, we introduce a hyperparameter $\lambda$ to scale the classification cross-entropy loss $\mathcal{L}_{ce}$. 
Accordingly, the total loss function of the model is formulated as follows:
\begin{equation}
\mathcal{L}=\lambda\mathcal{L}_{ce}+\mathcal{L}_{ss1} + \mathcal{L}_{ss2}
\label{concatloss}
\end{equation}
where $\mathcal{L}$ is the loss of one batch. Furthermore, we employ Mixup~\cite{mixup}, a data augmentation algorithm, to increase the quantity of the original data and improve the generalization ability and robustness of the model. Mixup is described in detail in the previous study~\cite{mixup}.

\begin{algorithm}[] \small  
	\caption{Pseudocode of the proposed RaGSECo.}\label{pseudocode}
    \label{Pseudocode}
	\LinesNumbered 
	\KwIn{$N_{epochs}$:~maximum epochs; $N_{batches}$:~total batchs; $\lambda$: hyper-parameter; $\mathcal{T}_{r}$: the set of training DDIs; $\bf{\hat{Y}}$:~ground truth of training DDIs; $\mathcal{T}_{e}$: the set of test DDIs.}
	\KwOut{$\bf{Y}_{te}$: predicted types of test DDIs.}
    $\bf{X} \leftarrow$ Obtain drugs’ initial features via (\ref{jaccard}). \\
    $\mathcal{G} \leftarrow$ Construct a DDI graph using known DDIs and $\bf{X}$. \\
    {//$\ast$ \textit{Training Procedure} $\ast$//} \\
    \For{ $ epoch \in [1,N_{epochs}]$}{
 	\For{ $ i \in [1,N_{batches}]$}{
    {//$\ast$ \textit{Acquiring Drug Embddings }$\ast$//} \\
    Use $\mathcal{G}$ learn RaGSEs of known drugs, ${\bf H}$, via Eq.(\ref{sub}).\\
    Use Jaccard similarities and ${\bf H}$ to build a DDS graph $\mathcal{G}_{s}$.\\
    Use $\mathcal{G}_{s}$ obtain RaGSEs of all drugs, ${\bf H}_{embed}$, via Eq.(\ref{propa1})-(\ref{propa4}).\\
    {//$\ast$ \textit{Acquiring Drug-Pair Representations }$\ast$//} \\
    Obtain $i$-th batch training DDIs from $\mathcal{T}_{r}$, and let the DP $k$ belong to the $i$-th batch training DDIs.\\
    Acquire three representations of DP $k$: ${\bf{p}}^{initial}_{k}$, ${\bf{p}}^{embed}_{k}$, ${\bf{S}}^{smile}_{k}$, via Eq.(\ref{views1})-(\ref{views3}).\\
    Obtain positive and negative sample set via Eq.(\ref{cosine}).\\
    $\mathcal{L}_{ss1}$, $\mathcal{L}_{ss2}$ $\leftarrow$ obtain co-contrastive learning losses via Eq.(\ref{coss1}) and Eq.(\ref{coss2}).\\
    Calculate final representations of DP $k$, ${\bf{p}}_{k}$, via Eq.(\ref{dpfinal1}) and Eq.(\ref{dpfinal2}).\\
    {//$\ast$ \textit{DDI Prediction }$\ast$//}\\
    Use ${\bf{p}}_{k}$ to calculate the probability distribution ${\bf{y}}_{k}$ via Eq.(\ref{predict}).\\
    $\mathcal{L}_{ce} \leftarrow$ use ground truth $\bf{\hat{y}_{k}}$ and predicted result $\mathbf{y}_{k}$ calculate prediction loss via Eq.(\ref{lossce}).\\
    Obtain the final loss $\mathcal{L}\leftarrow \lambda\mathcal{L}_{ce} + \mathcal{L}_{ss1} + \mathcal{L}_{ss2}$.\\
    Minimize $\mathcal{L}$ and update ${\bf H}$, ${\bf H}_{embed}$, and ${\bf{p}}_{k}$.\\
	}
	}
    {//$\ast$ \textit{Testing Procedure} $\ast$//} \\
    Predict types of the DDI in $\mathcal{T}_{e}$ using the trained MSAGC model.\\
\end{algorithm}

RaGSECo is a semi-supervised DDI prediction approach.
Both known and new drugs are incorporated into the training and testing phases of each task.
We present pseudocode in Algorithm \ref{Pseudocode} for a clearer illustration of optimizing the stated objective. 
We begin by extracting drug features via Jaccard similarities (Line 1).
Following this, we construct a multi-relational DDI graph using the extracted drug features and known DDIs, derived from training data $\mathcal{T}_{r}$ (Line 2). 
During each batch of each training iteration, our initial step is to learn embeddings ${\bf H}_{embed}$ for all drugs (Lines 7-9).
Then, a batch of training data is obtained from $\mathcal{T}_{r}$, and feature representations ${\bf{P}}_{k}$ for the corresponding DPs are learned (Lines 11, 12, and 15).
Subsequently, we construct sets of positive and negative samples (Line 16) and calculate co-contrastive learning losses $\mathcal{L}_{ss1}$, $\mathcal{L}_{ss2}$ (Line 13 and 14). 
The prediction loss, denoted as $\mathcal{L}_{ce}$, is subsequently computed (Lines 17 and 18).
In the following step, we merge $\mathcal{L}_{ss1}$, $\mathcal{L}_{ss2}$ and $\mathcal{L}_{ce}$ to generate the final loss $\mathcal{L}$ (Line 19), then optimize $\mathcal{L}$ to update ${\bf H}_{embed}$ and ${\bf{P}}_{k}$ (Line 20).
Once all training iterations are complete, we use the trained RaGSECo to test the types of interaction events of DPs in $\mathcal{T}_{e}$.

\section{Results and Discussion}\label{results}

\subsection{Datasets}\label{dataset}
This paper uses two real datasets to explore the effectiveness and competitiveness of our RaGSECo. The first data set (Dataset 1) was collected by Deng et al. from DrugBank~\footnote{\url{https://go.drugbank.com/releases/latest}} and published in~\cite{DDIMDL}.
Dataset 1 contains $572$ drugs with $37~264$ pairwise DDIs associated with $65$ interaction types. 
Each drug is represented by four biological attributes: enzymes, targets, pathways, and chemical substructures.

We also construct a dataset (Dataset 2) from DrugBank. 
We use $1000$ drugs, and each drug is described by three features: enzymes, targets, and chemical substructures. 
Accordingly, we obtained a total of $206~029$ pairwise DDIs, which are associated with $99$ types of events. 
The detailed information on the two datasets is listed in Table~\ref{table1}.

\begin{table}
\caption{Summary of the datasets utilized in our experiments.}%
\begin{tabular*}{\columnwidth}{@{\extracolsep\fill}llll@{\extracolsep\fill}}
\toprule
Dataset & Drug number  & DDI number  & DDI event types \\
\midrule
Dataset 1    & 572  & 37264  & 65  \\
Dataset 2    & 1000 & 206029  & 99 \\
\bottomrule
\end{tabular*}
\label{table1}
\end{table}

\subsection{Baselines}\label{Baselines}
We compare the proposed RaGSECo with six baselines:

$\bullet$ MCFF-MTDDI~\cite{MCFF-MTDDI} extracts multiple drug-related features and proposes a gated recurrent unit-based multi-channel feature fusion module to yield comprehensive representations of DPs.

$\bullet$ MDF-SA-DDI~\cite{MDF-SA-DDI} uses four encoders to generate four different DP features and adopts transformers to perform feature fusion.

$\bullet$ RANEDDI~\cite{RANEDDI} is a relation-aware graph structure embedding method that considers the multi-relational interaction information to obtain drug embedding for DDI prediction. 

$\bullet$ DDIMDL~\cite{DDIMDL} is a DNN method that pays attention to multiply drug-drug-similarities for multi-relational DDI prediction. 

$\bullet$ DeepDDI~\cite{DeepDDI} is a DNN method that uses structural information of DPs to train the model and make DDI predictions.

$\bullet$ DNN~\cite{DNN} uses multiple similarity information of DPs to predict the pharmacological effects of DDIs by autoencoders and a deep feed-forward network.

\subsection{Experimental Settings}\label{experisett}
To demonstrate the effectiveness, we evaluate our RaGSECo by implementing Tasks 1, 2, and 3 on the two real datasets. 
In Task 1, we randomly split DDIs into five subsets via 5-fold cross-validation (5-CV), with four subsets as the training set and the remaining as the test set.
For Tasks 2 and 3, we randomly split drugs instead of DDIs into five subsets via 5-CV, with four subsets used as the known drugs and the remaining as the new drugs.
In Task 2, the DDIs involving two known drugs are defined as the training set, and the DDIs involving one known drug and one new drug are utilized as the test set.
In Task 3, the training set is the same as Task 2, and the DDIs between two new drugs are utilized as the test set.

The prediction task is the multi-class classification work.
To evaluate the classification performance of our RaGSECo,
we use six metrics: accuracy (ACC), the area under the precision–recall-curve (AUPR), the area under the ROC curve (AUC), Precision, Recall, and F1 score.
The activation function, dropout layer, and batch normalization layer \cite{Batchnorma} are used between the fully connected layers. 
By default, we use the Gaussian error linear unit \cite{gelu} activation function and Radam optimizer \cite{Radam}. 
The proposed RaGSECo is implemented with the deep learning library PyTorch.
The Python and PyTorch versions are 3.8.10 and 1.9.0, respectively.
All experiments are conducted on a Windows server with a GPU (NVIDIA GeForce RTX 4090 Ti).

\subsection{Hyperparameter Searching and Setting}\label{Hyperparameterss}

The proposed RaGSECo involves some hyperparameters that influence the prediction performance, including batch size ($bs$), learning rate ($lr$), dropout rate ($dr$), training epochs ($te$), the dimension of drug RaGSEs ($d'$), the power of adjacency matrices in the DDS graph ($n$), the output dimension of FNN1 ($d^{FNN}$), threshold values ($t_{pos}$ and $t_{neg}$), and the weight of CE loss ($\lambda$).
To thoroughly investigate the impact of each parameter on the prediction results, a grid search strategy is employed where one parameter is varied while keeping the other parameters fixed.
Table~\ref{hyperparameters_value} summarizes the values of all hyperparameters on different datasets and tasks.
Among these hyperparameters, $n$, $d^{FNN}$, $t_{pos}$ and $t_{neg}$ are specific to the proposed RaGSECo.
Therefore, in the subsequent experiments, we will provide the impact of these hyperparameters on the experimental results and analyze the reasons for them.

\subsubsection{Impact of RaGSE Propagation Distance $n$}
\begin{figure*}[t]
\centering
\includegraphics[width=0.55in,height=1.3in]{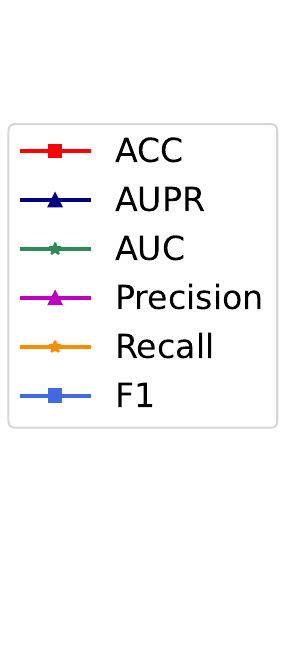}
\subfigure[]{\includegraphics[width=1.35in,height=1.3in]{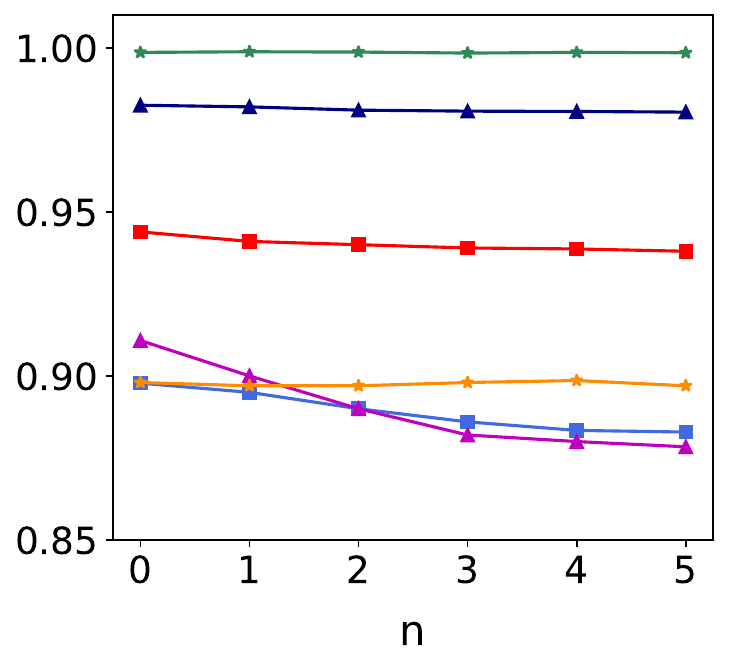}}
\subfigure[]{\includegraphics[width=1.35in,height=1.3in]{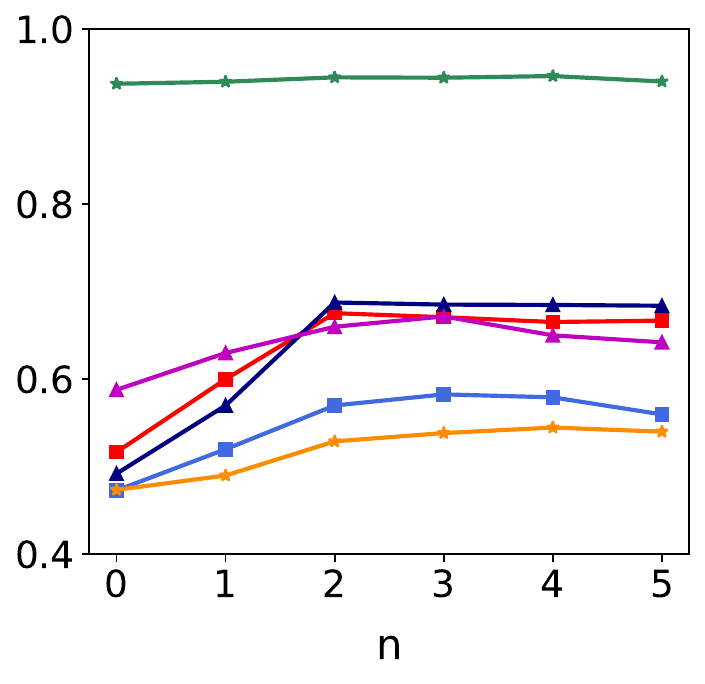}}
\subfigure[]{\includegraphics[width=1.35in,height=1.3in]{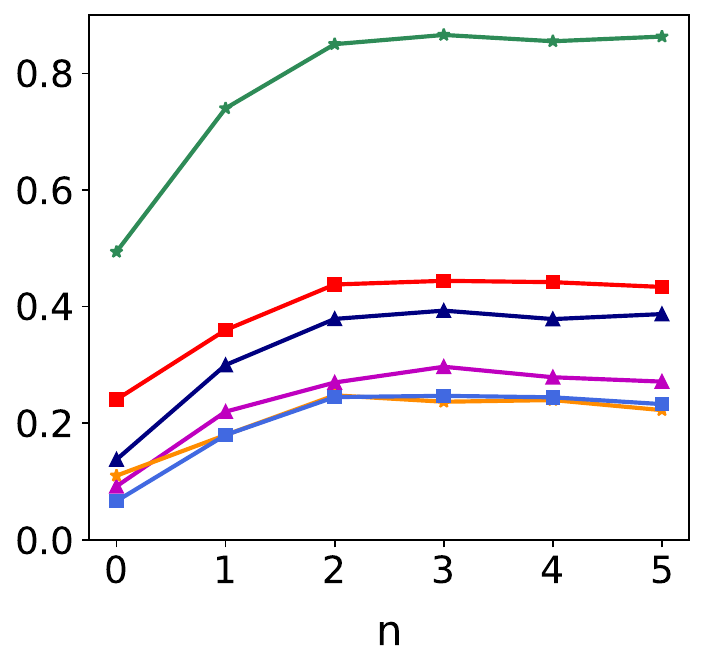}}
\caption{Six metric scores versus the maximum RaGSE propagation distance $n$ on Task 1 (a), Task 2 (b), Task 3 (c) of Dataset 1.}
\label{n-hop}
\end{figure*}
The hyperparameter $n$ represents the power of adjacency matrices in the DDS graph and determines the distance of RaGSE propagation. 
It plays a crucial role in RaGSE propagation.
To understand the influence of $n$ on the prediction performance, we conduct experiments on three tasks of Dataset 1 and evaluate six metric scores across different $n$ values.
Referring to Fig.~\ref{n-hop} (a), in Task 1, the metric scores of RaGSECo do not change drastically as $n$ increases, and RaGSECo can consistently achieve good prediction performances.
On the other hand, in Tasks 2 and 3, we observed that the prediction performance of RaGSECo is sensitive to the value of $n$.
Specifically, when $n$ is set to 0, the prediction performance of our RaGSECo is at its lowest.
The prediction performance significantly improves as the value of $n$ increases. 
This is because the test DDIs in Tasks 2 and 3 involve new drugs that do not possess relation-aware interaction information when $n$ is 0. As a result, the model tends to be over-fitting.
As $n$ increases, new drugs can aggregate effective relation-aware interaction information from similar known drugs, which significantly improves the generalization ability of RaGSECo.
Based on the experimental results, we determined that setting $n$ as 0 in Task 1 and 3 in Tasks 2 and 3 leads to desirable prediction performances.

\subsubsection{Impact of Output Dimension $d^{FNN}$}
\begin{figure*}[t]
\centering
\includegraphics[width=0.55in,height=1.3in]{Hyperpara/legend.pdf}
\subfigure[]{\includegraphics[width=1.35in,height=1.3in]{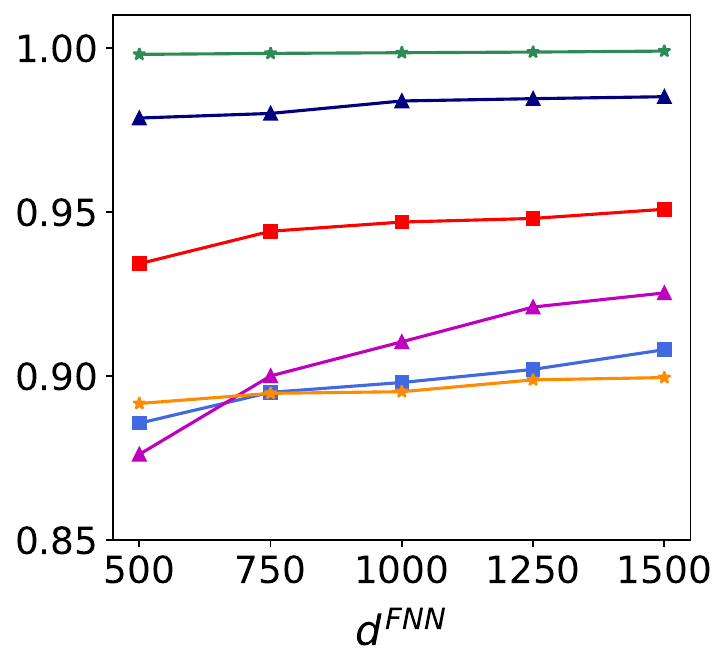}}
\subfigure[]{\includegraphics[width=1.35in,height=1.3in]{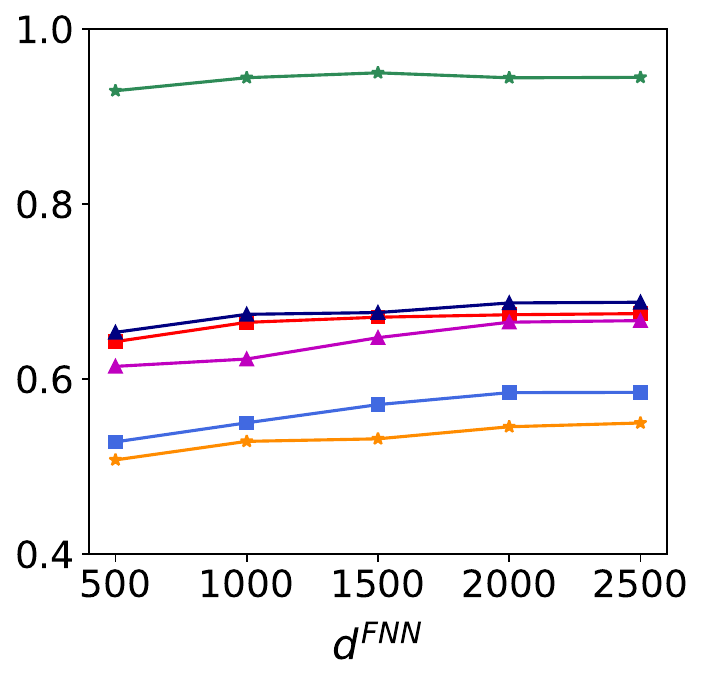}}
\subfigure[]{\includegraphics[width=1.35in,height=1.3in]{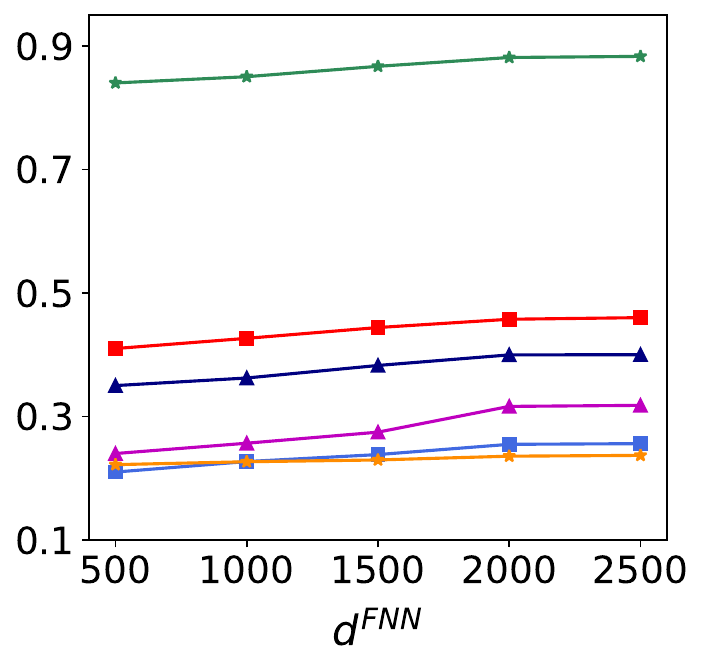}}
\caption{Six metric scores versus the output dimension of FNN1 and FNN2 $d^{FNN}$ on Task 1 (a), Task 2 (b), and Task 3 (c) of Dataset 1.}
\label{dAE}
\end{figure*}

The output dimension $d^{FNN}$ of FNN1 and FNN2 is an important hyperparameter for our RaGSECo model. Increasing $d^{FNN}$ can enhance the generalization ability of RaGSECo to some extent.
In order to investigate the impact of $d^{FNN}$ on the prediction performance of RaGSECo, we conducted experiments on three tasks of Dataset 1 and examined the changes in metric scores as $d^{FNN}$ varies.
From Fig.~\ref{dAE}, it can be observed that the prediction performances of RaGSECo gradually improve as $d^{FNN}$ increase among all tasks. 
This indicates the robustness of our model.
To strike a balance between accuracy and efficiency, we set $d^{FNN}$ to 1000 for Task 1 and 1500 for Tasks 2 and 3.

\subsubsection{Impact of Threshold Values $t_{pos}$ and $t_{neg}$}
\label{ImpactoH}

\begin{figure*}[h]
\centering
\subfigure[]{\includegraphics[width=1.5in,height=1.2in]{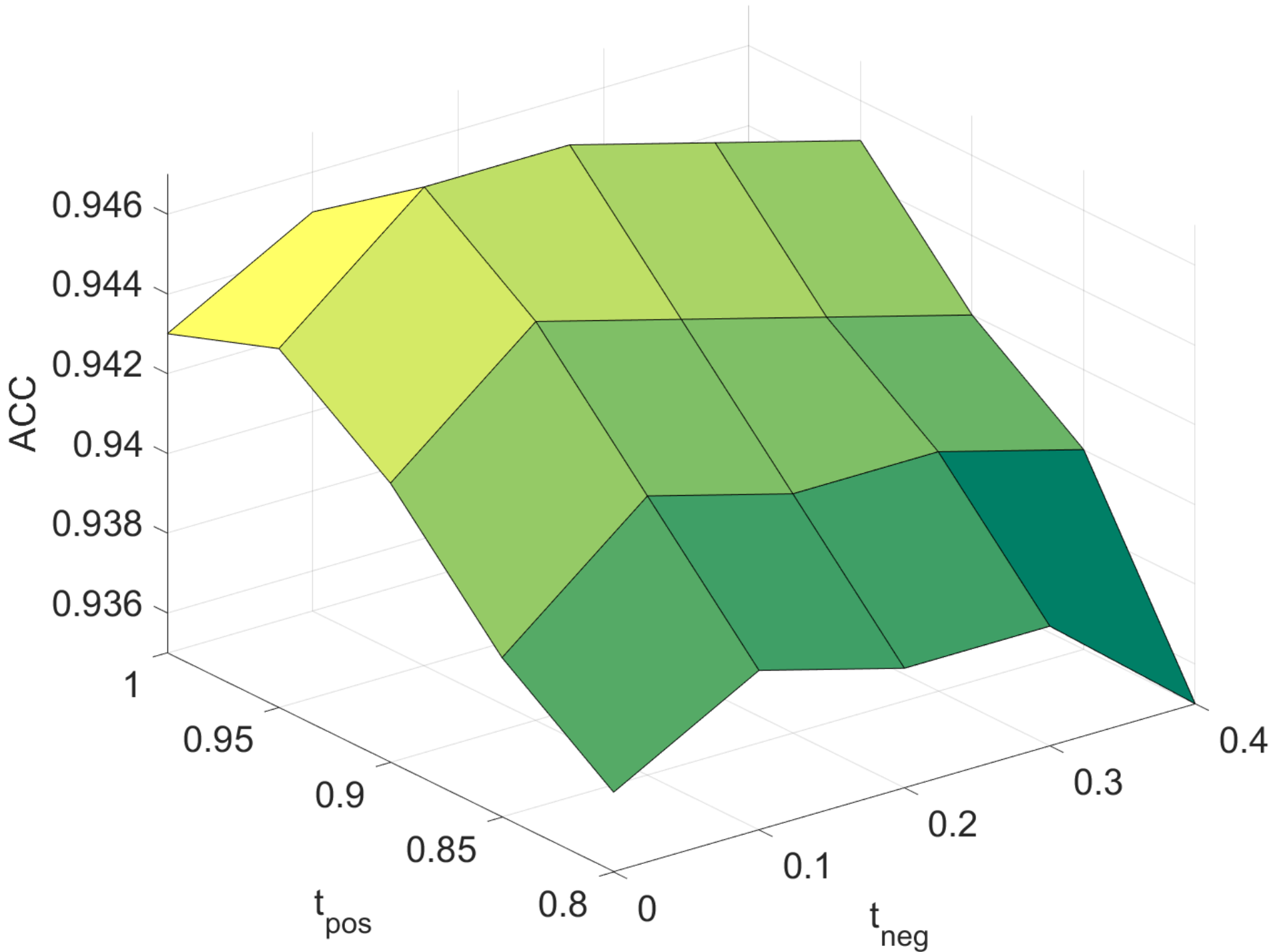}}\hspace{-1mm}
\subfigure[]{\includegraphics[width=1.5in,height=1.2in]{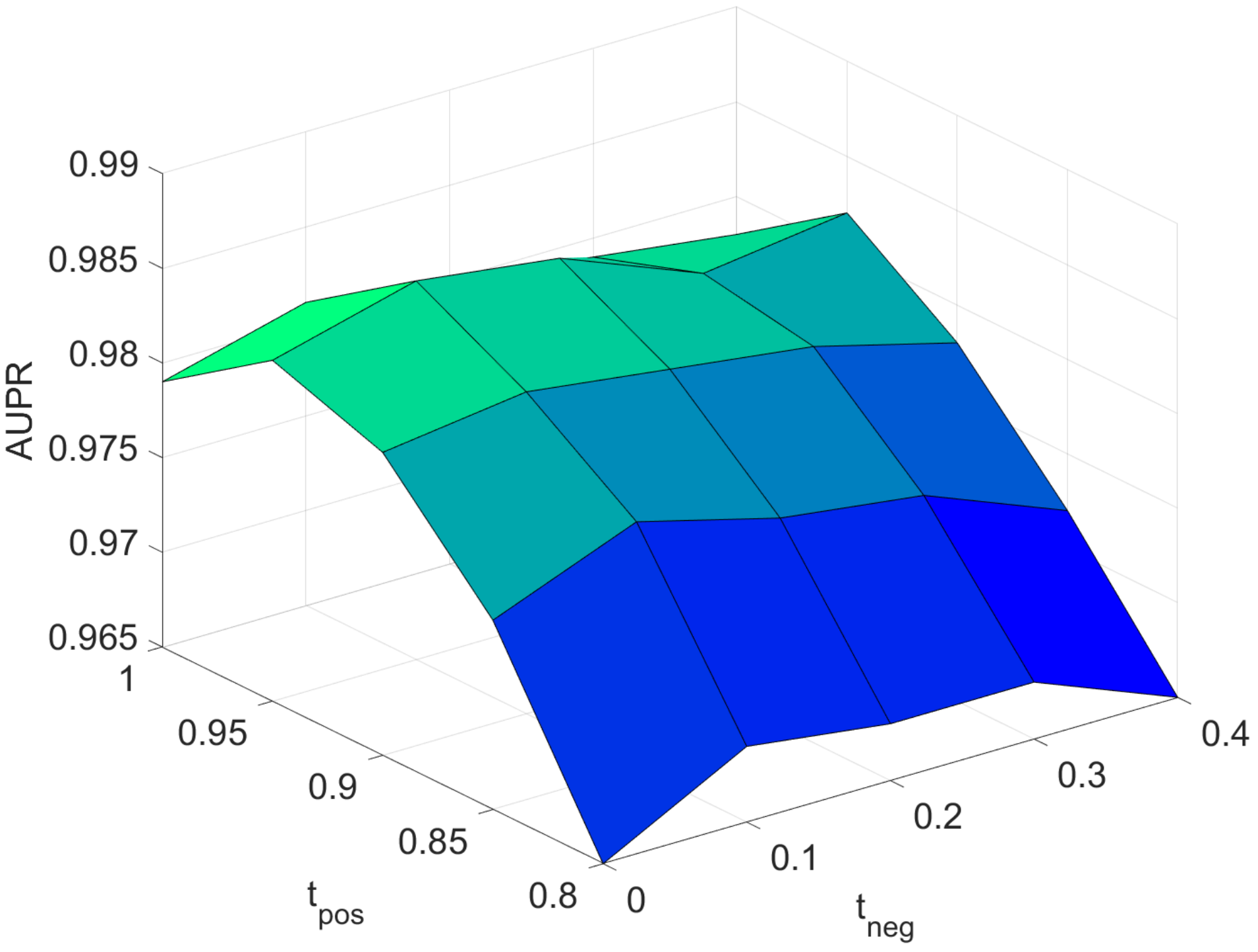}}\hspace{-1mm}
\subfigure[]{\includegraphics[width=1.5in,height=1.2in]{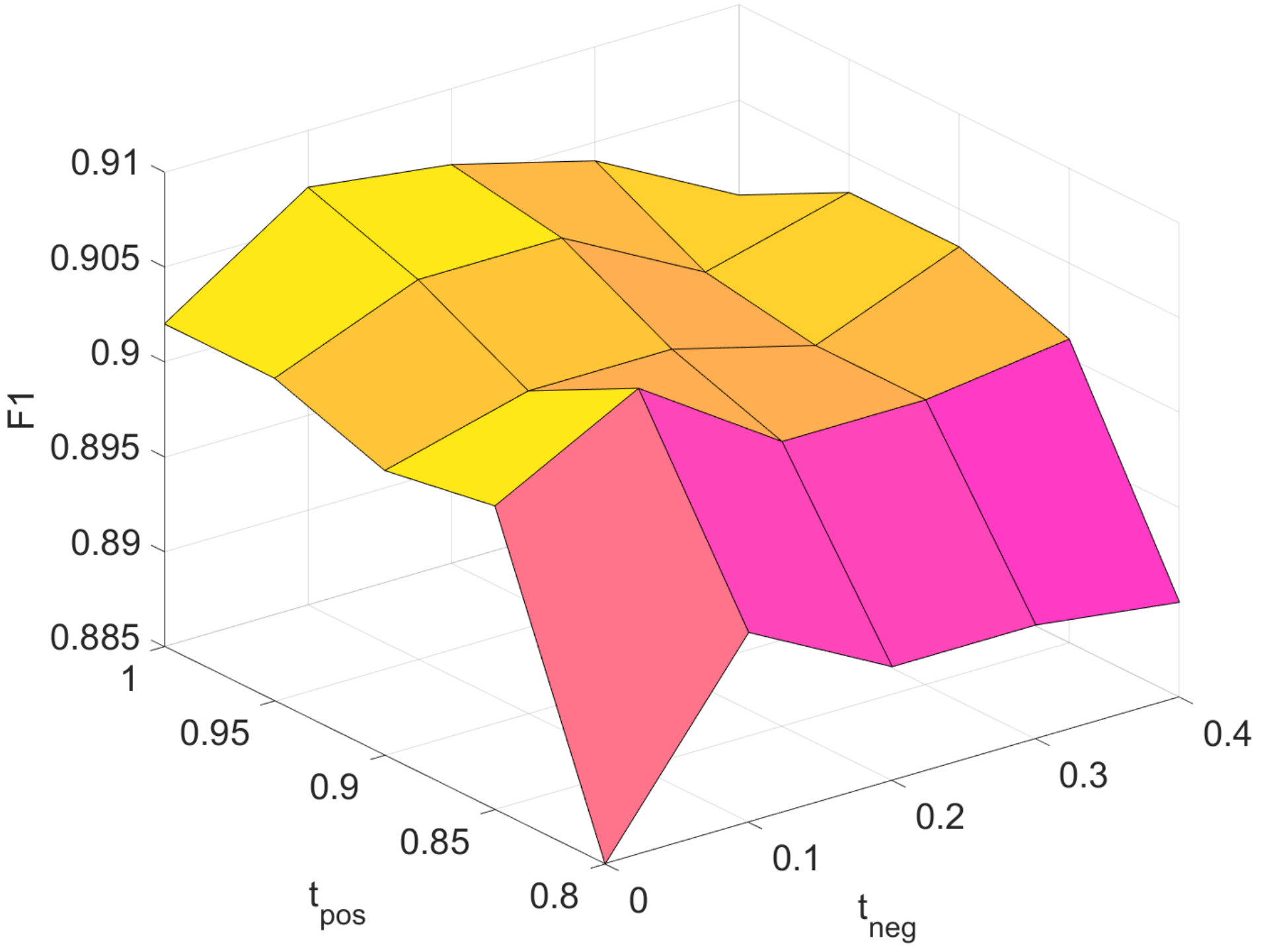}}\hspace{-1mm}
\caption{ACC (a), AUPR (b), and F1(c) versus the threshold values $t_{pos}$ and $t_{neg}$ on Task 1 of Dataset 1.}
\label{dtt}
\end{figure*}

The threshold values $t_{pos}$ and $t_{neg}$ are important parameters for selecting positive and negative samples of DPs.
In our experiments, we set $t_{pos}$ to the range of $[0.8, 0.85, \dots, 1]$ and $t_{neg}$ to the range of $[0, 0.1, \dots, 0.4]$.
Fig. \ref{dtt} presents the comparisons of three representative metric scores (ACC, AURR, and F1) for different values of $t_{pos}$ and $t_{neg}$.
It can be observed that the results are not significantly affected by variations in $t_{pos}$ and $t_{neg}$.
This indicates the stability of our co-contrastive learning strategy. 
Herein, the feature representations from the two views of each DP serve as positive samples for each other when $t_{pos}$ equals 1.
As observed, the model produces superior prediction results when $t_{pos}$ is set to 0.95, as opposed to when it is set to 1. 
This observation indicates the effectiveness of our positive sample selection strategy.
Finally, we set $t_{pos}=0.95$ and $t_{neg}=0.1$ for subsequent experiments.

\begin{figure*}[t]
\centering
\small
\subfigure[]{\includegraphics[width=1.5in,height=0.9in]{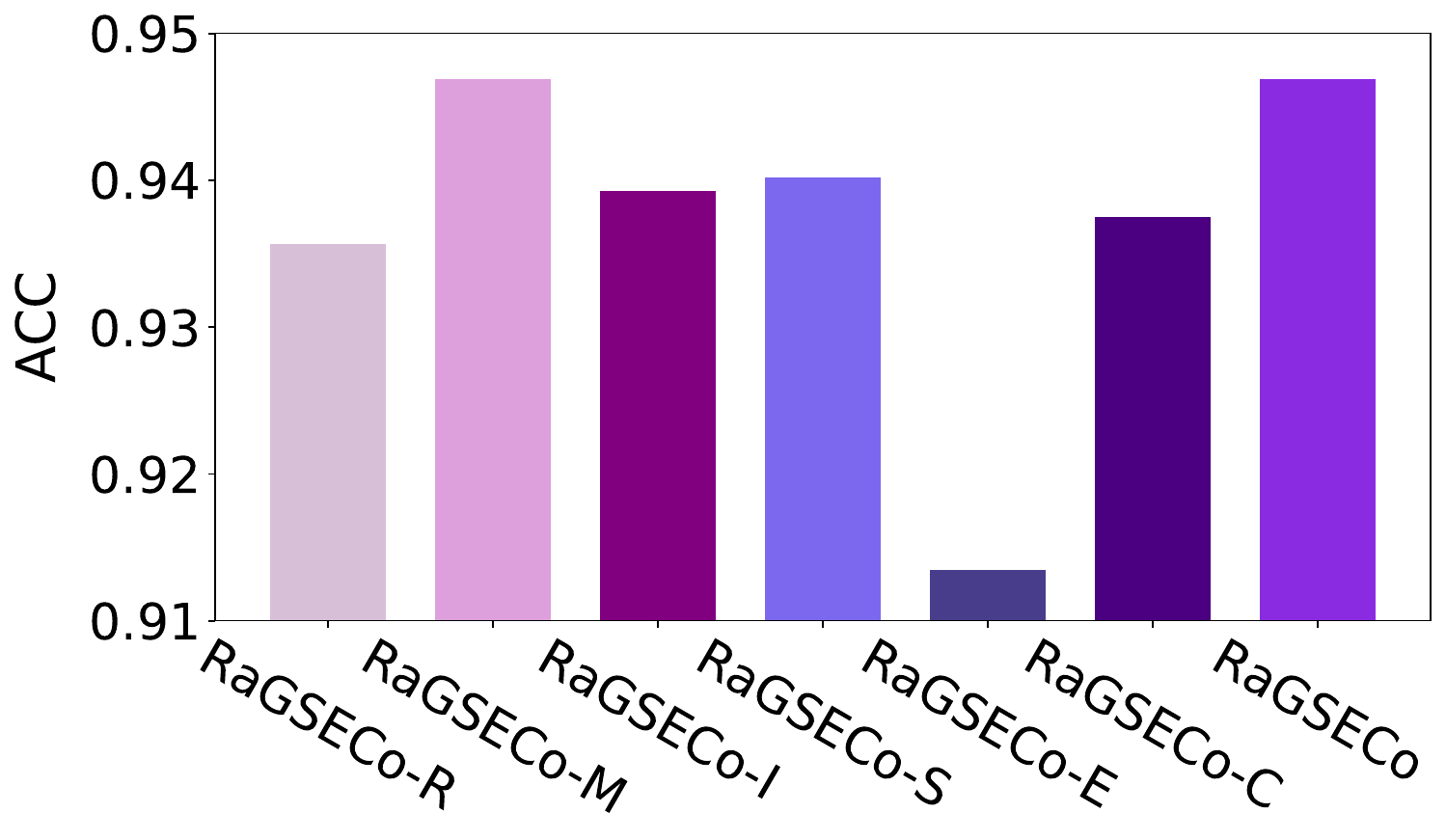}\label{variantsACC1}}
\subfigure[]{\includegraphics[width=1.5in,height=0.9in]{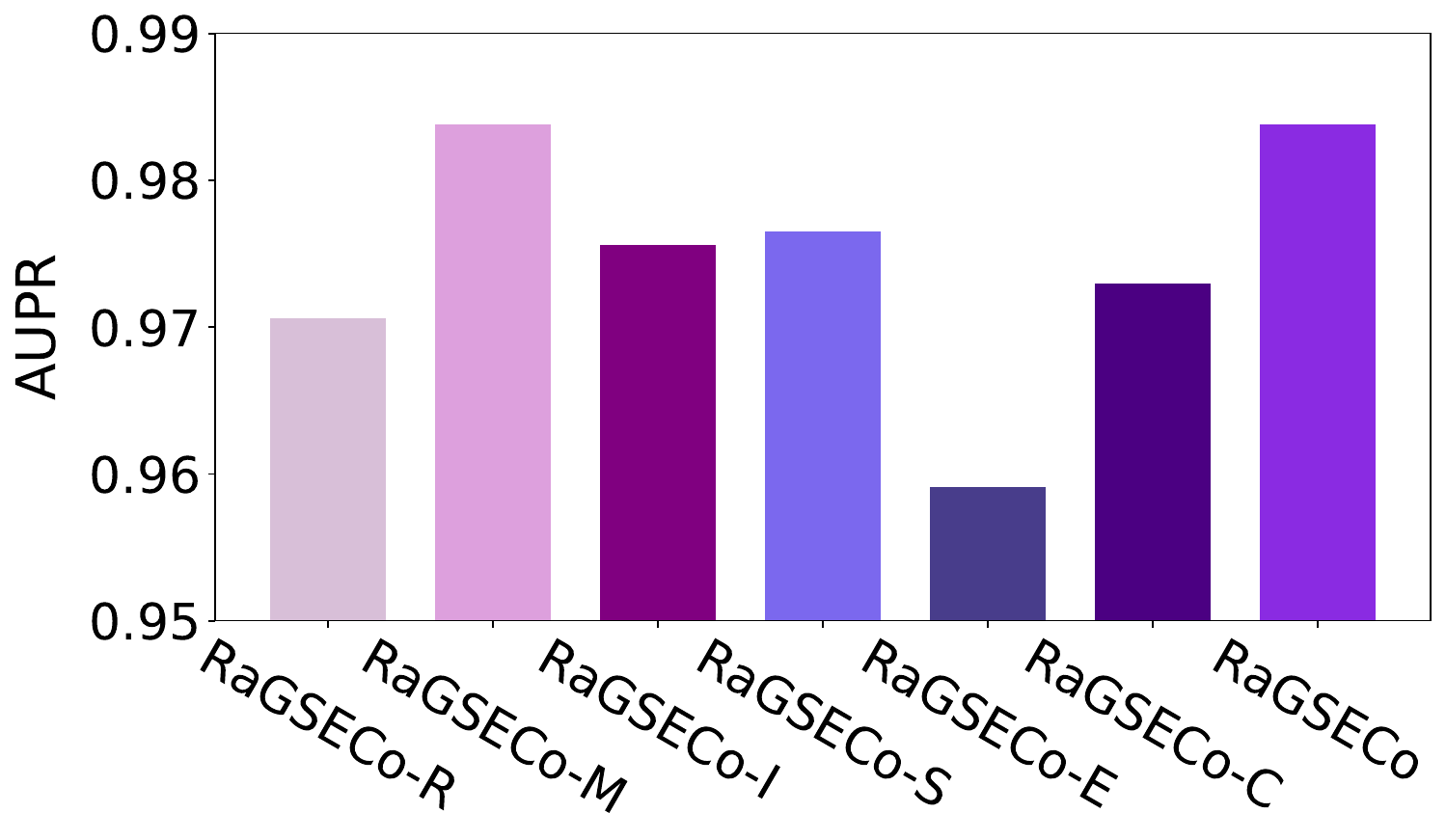}\label{variantsAUPR1}}
\subfigure[]{\includegraphics[width=1.5in,height=0.9in]{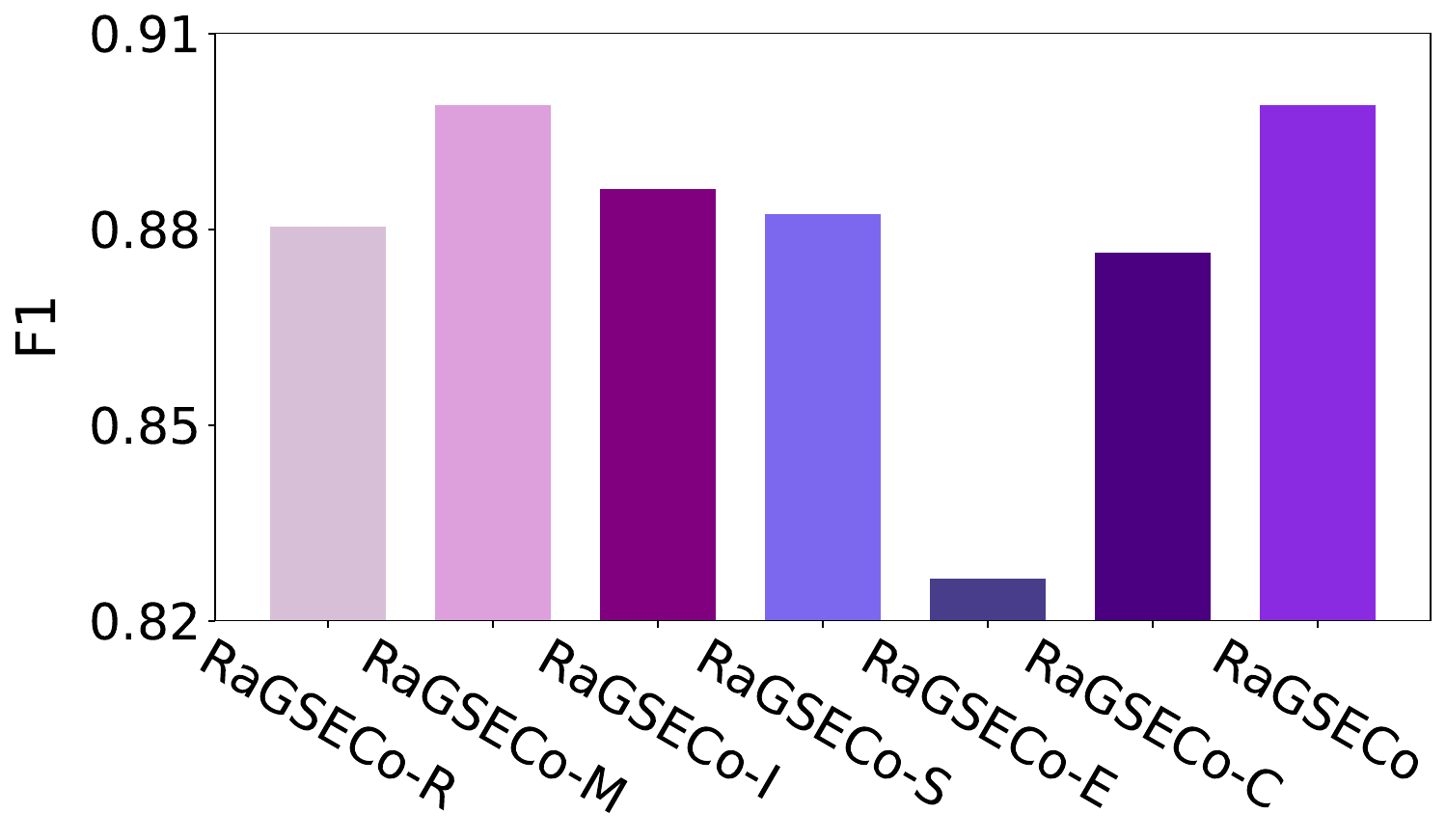}\label{variantsF11}}
\caption{Experimental results of RaGSECo and its six variants in terms of ACC (a), AUPR (b), and F1 (c) on Task 1 of Dataset 1.}
\label{variants1}
\end{figure*}

\begin{figure*}[t]
\centering
\subfigure[]{\includegraphics[width=1.5in,height=0.9in]{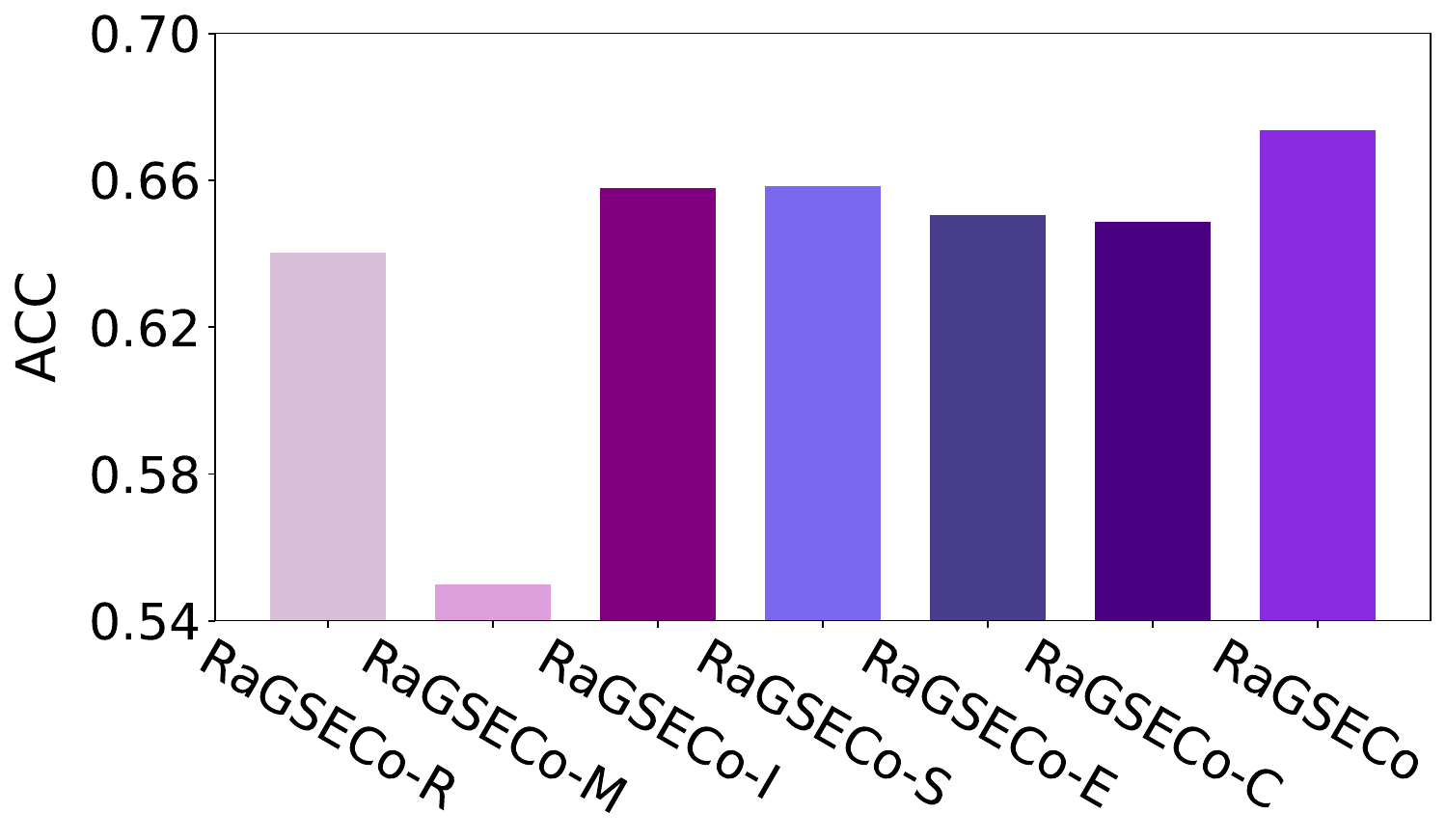}}
\subfigure[]{\includegraphics[width=1.5in,height=0.9in]{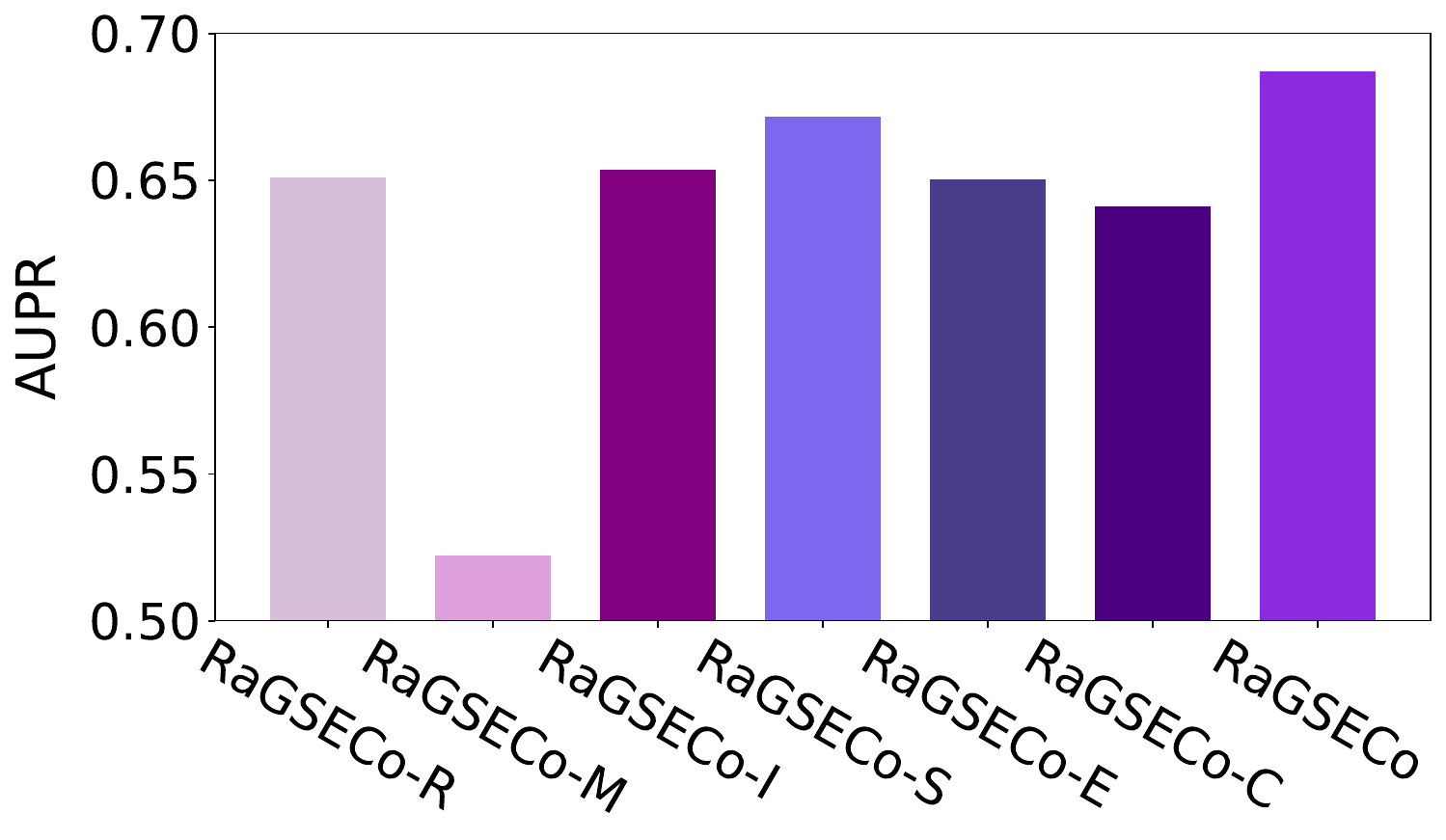}} 
\subfigure[]{\includegraphics[width=1.5in,height=0.9in]{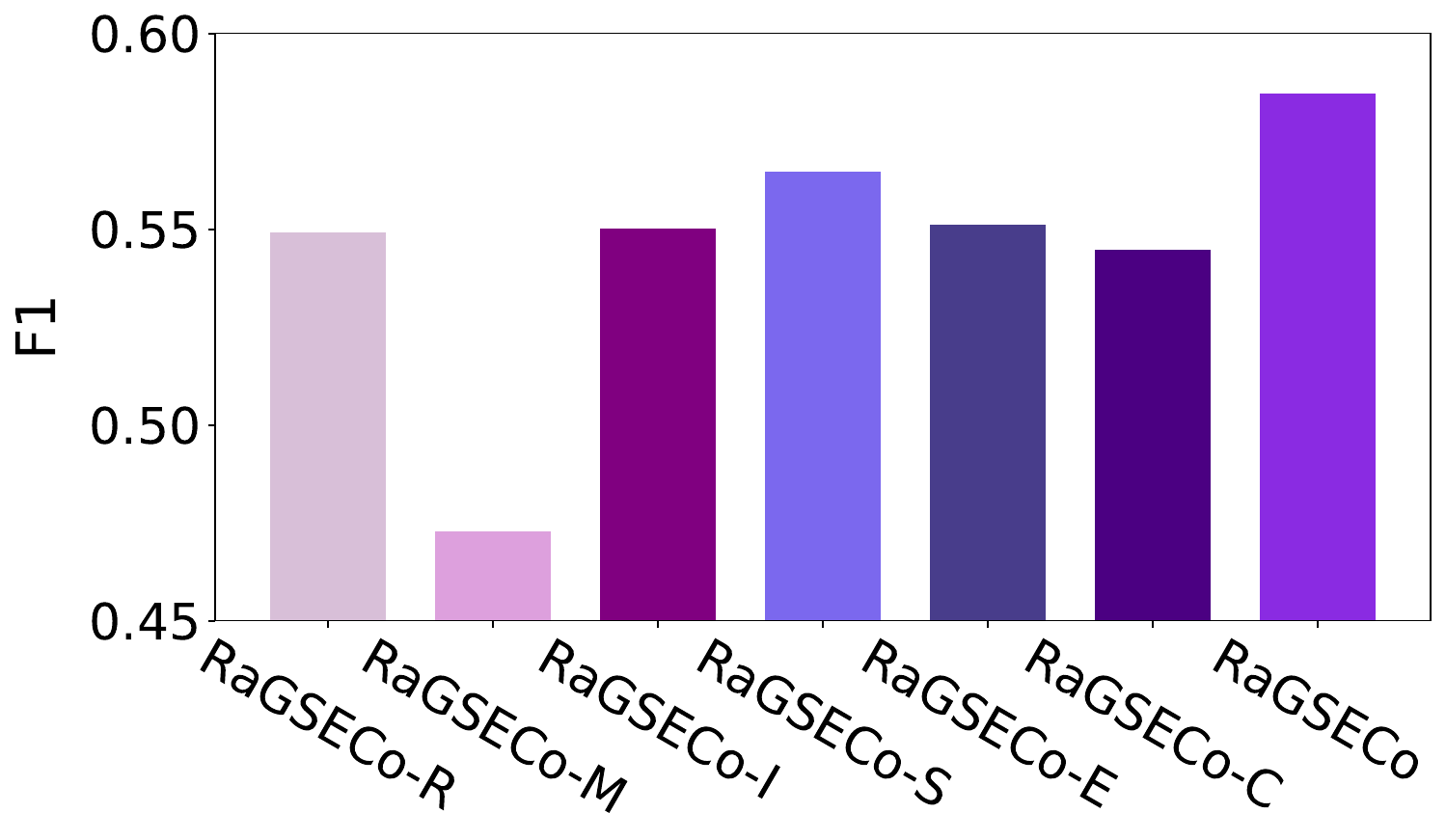}}
\caption{Experimental results of RaGSECo and its six variants in terms of ACC (a), AUPR (b), and F1 (c) on Task 2 of Dataset 1.}
\label{variants2}
\end{figure*}
\begin{figure*}[t]
\centering
\subfigure[]{\includegraphics[width=1.5in,height=0.9in]{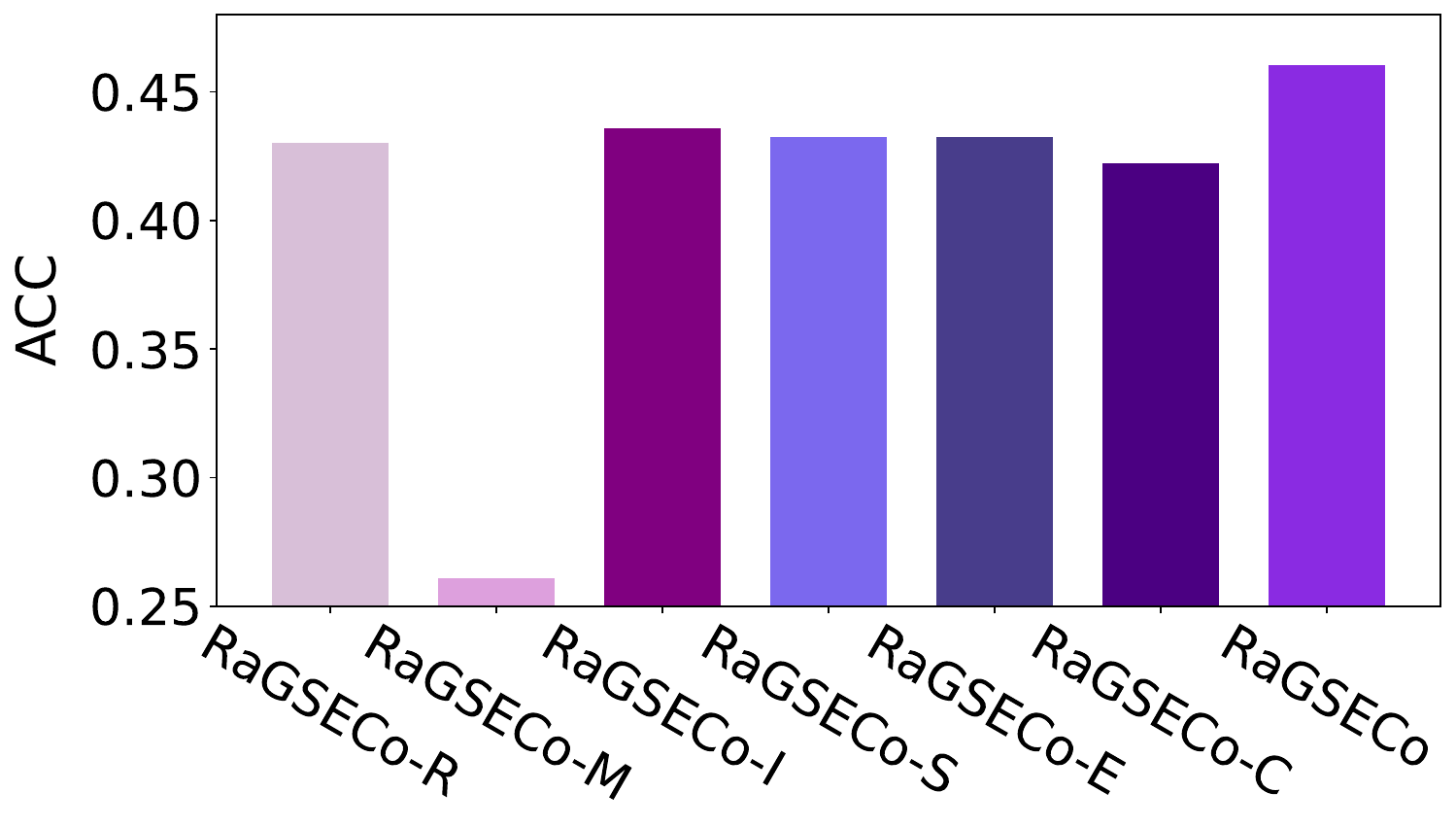}}
\subfigure[]{\includegraphics[width=1.5in,height=0.9in]{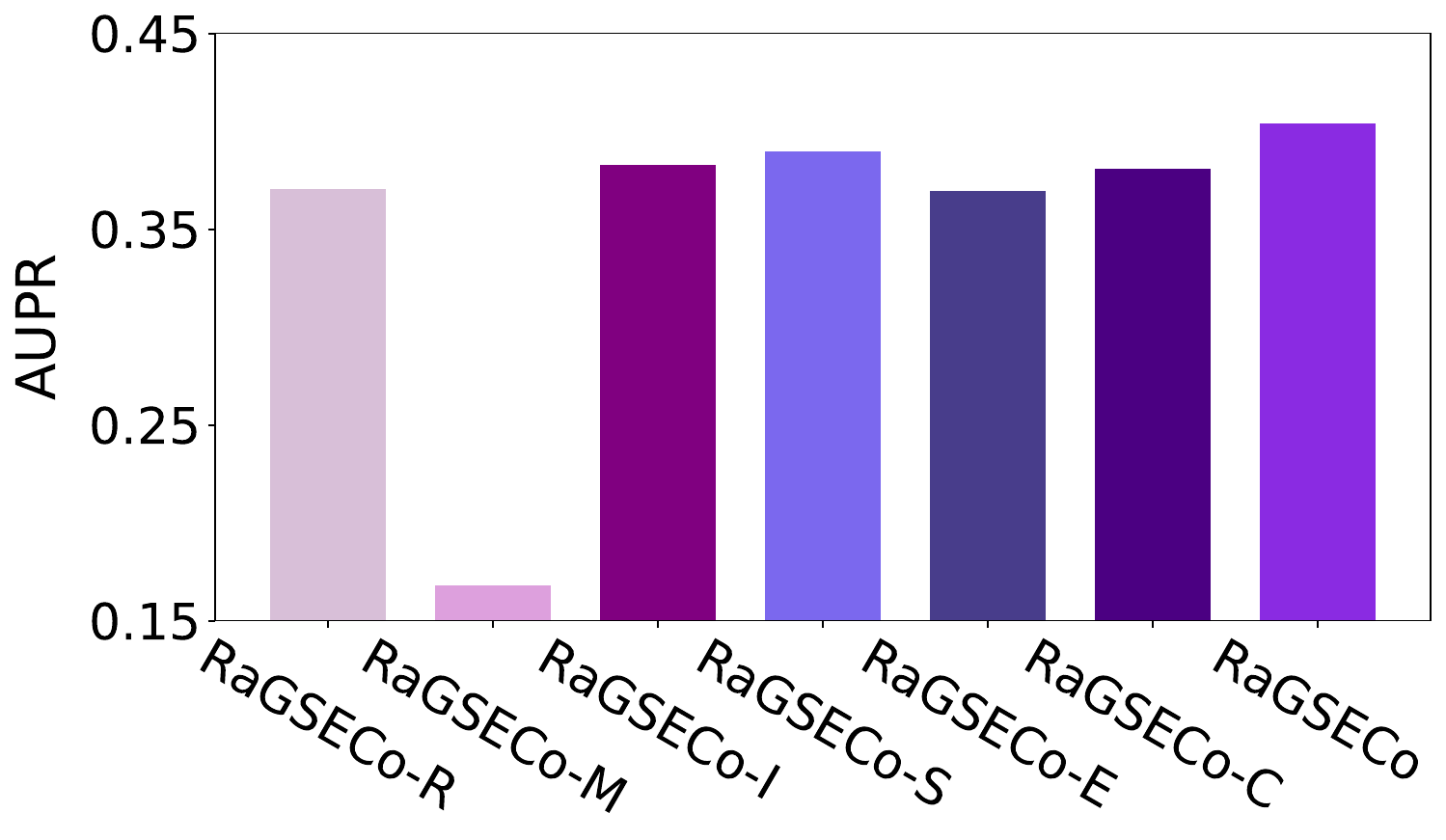}}
\subfigure[]{\includegraphics[width=1.5in,height=0.9in]{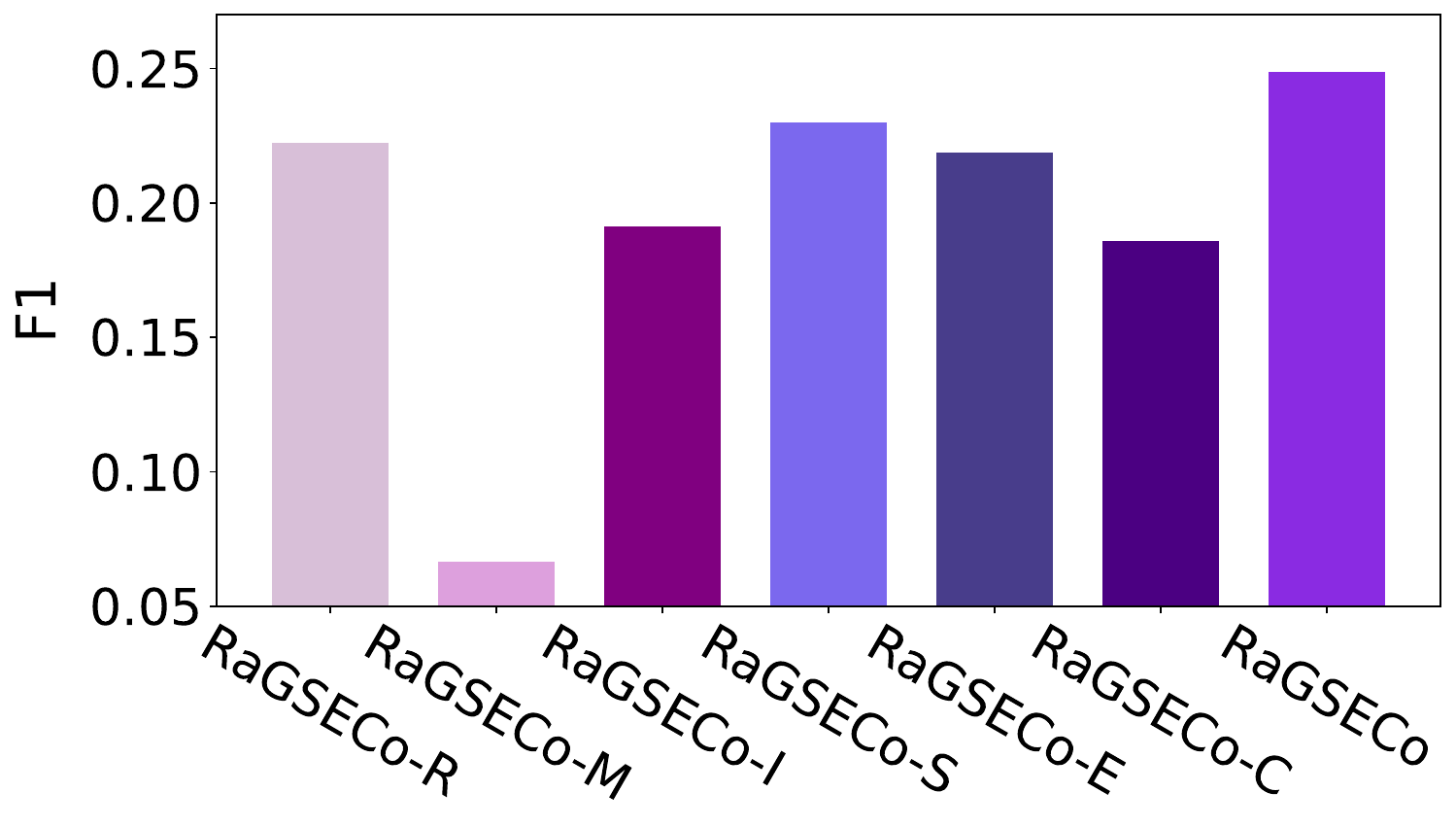}}
\caption{Experimental results of RaGSECo and its six variants in terms of ACC (a), AUPR (b), and F1 (c) on Task 3 of Dataset 1.}
\label{variants3}
\end{figure*}

\subsection{Analysis of RaGSECo with Its Variants}\label{Variants}

In this subsection, we compare our proposed RaGSECo with six variants to assess the necessity and effectiveness of each component in DDI prediction. RaGSECo incorporates multiple components, including the construction of multi-relational DDI and multi-attribute DDS graphs, the integration of SMILES string, initial features, and RaGSEs of drugs, and the adoption of a co-contrastive learning mechanism to enhance DP representation learning.
By comparing RaGSECo with its variants, we aim to understand the contributions of each component to the overall performance. The six variants are represented as follows:

$\bullet~$RaGSECo-R: A variant of RaGSECo that does not construct a multi-relational DDI graph to learn RaGSEs of known drugs. From a practical perspective, the initial features of drugs $\mathbf{X}$ are directly used as the  node initial features of the multi-attribute DDS graph to learn drug embeddings.

$\bullet~$RaGSECo-M: A variant of RaGSECo that does not construct the multi-attribute DDS graph for RaGSE propagation. In other words, the output of RaGSEL, $\mathbf{H}$, is taken as the final drug embeddings.

$\bullet~$RaGSECo-I: A variant of RaGSECo that neglects the initial features of drugs during DP representation learning. Hence, the representations of DP $k$ in Eq.~(\ref{dpfinal2}) are denoted as:
$\bf{p}_{k} =||\left({\bf p}_{k}^{smile},{\bf p}_{k}^{embed},{\bf p}_{k}^{smile}+{\bf p}_{k}^{embed}\right)$. 

$\bullet~$RaGSECo-S: A variant of RaGSECo that does not use the SMILES strings of drugs during DP representation learning. Thus, the representations of DP $k$ in Eq.~(\ref{dpfinal2}) are represented as:
$\bf{p}_{k} =||\left( {\bf p}_{k}^{initi}, {\bf p}_{k}^{embed}, {\bf p}_{k}^{initi}+ {\bf p}_{k}^{embed}\right)$.

$\bullet~$RaGSECo-E: A RaGSECo variant that ignores drug RaGSEs during DP representation learning. Hence, the representations of DP $k$ in Eq.~(\ref{dpfinal2}) are denoted as:
$\bf{p}_{k} =||\left({\bf p}_{k}^{initi}, {\bf p}_{k}^{smile}, {\bf p}_{k}^{initi}+ {\bf p}_{k}^{smile}\right)$.  

$\bullet~$RaGSECo-C: A RaGSECo variant that neglects the employment of co-contrastive learning. 
Hence, the representations of DP $k$ in Eq.~(\ref{dpfinal2}) are denoted as: $\bf{p}_{k} ={\bf p}_{k}^{embed}$.
the total loss is represented as $\mathcal{L} = \mathcal{L}_{ce}$.

Herein, we select three representative metric scores (Accuracy, AUPR, and F1) to evaluate the prediction performance of these models.
Fig. \ref{variants1}, \ref{variants2}, and \ref{variants3} illustrate the performance of RaGSECo and its six variants on Task 1, 2, and 3 of Dataset 1, respectively. 
These figures show that RaGSECo achieves higher metric scores compared to its variants, indicating the effectiveness of RaGSE learning and propagation, multimodal DP representation learning, and the co-contrastive learning mechanism.
Comparing RaGSECo-R and RaGSECo-M, we can see that RaGSECo-M performs better on Task 1, while RaGSECo-R significantly outperforms RaGSECo-M on Tasks 2 and 3. This observation confirms that the test DDIs in Task 1 consist of known drugs with distinguishable relation-aware information.
Meanwhile, it demonstrates that the test DDIs include new drugs in Tasks 2 and 3, which may impact the model performance.  
Nevertheless, RaGSE propagation can effectively mitigate this issue.
Analyzing the results of RaGSECo-I, RaGSECo-S, and RaGSECo-E reveals that incorporating the drug's initial features, SMILES string, and RaGSEs enhances data diversity and improves the model's performance.
Finally, the performance of RaGSECo-C further validates the effectiveness of the co-contrastive learning mechanism.

\begin{table*}[t]\small
\caption{Prediction performances of different methods on Dataset 1.}
\tabcolsep=0pt
\begin{tabular*}{\textwidth}{@{\extracolsep{\fill}}lccccccc@{\extracolsep{\fill}}}
\toprule
& Method $~$ & ACC $~$&  AUPR $~$ & AUC$~$ & Precision$~$ & Recall & F1$~$\\
\midrule
\multirow{7}{*}{Task$~$1}
& RaGSECo       & \bf{0.9461} & \bf{0.9838} & \bf{0.9991}  & \bf{0.9121} & \bf{0.9043} & \bf{0.9050} \\
& MCFF-MTDDI      & 0.9350 & 0.9757 & 0.9985 & 0.9100 & 0.8820 & 0.8918 \\ 
& MDF-SA-DDI      & 0.9301 & 0.9737 & 0.9989 & 0.9085 & 0.8760 & 0.8878 \\
& RANEDDI         & 0.9228 & 0.9657 & 0.9980 & 0.8747 & 0.8701 & 0.8717 \\
& DDIMDL          & 0.8852 & 0.9208 & 0.9976 & 0.8471 & 0.7182 & 0.7585  \\
& DeepDDI         & 0.8371 & 0.8899 & 0.9961 & 0.7275 & 0.6611 & 0.6848  \\
& DNN             & 0.8797 & 0.9134 & 0.9963 & 0.8047 & 0.7027 & 0.7223  \\
\midrule
\multirow{5}{*}{Task$~$2}
& RaGSECo          & \bf{0.6826} & \bf{0.7002} & 0.9535 & 0.6514 & \bf{0.5631} & \bf{0.5860}   \\
& MCFF-MTDDI      & 0.6650 & 0.6800 & 0.9500 & \bf{0.6561} & 0.5139 & 0.5574  \\ 
& MDF-SA-DDI      & 0.6633 & 0.6776 & 0.9497 & 0.6547 & 0.5078 & 0.5584   \\
& DDIMDL          & 0.6415 & 0.6558 & \bf{0.9799} & 0.5607 & 0.4319  & 0.4460 \\
& DeepDDI         & 0.5774 & 0.5594 & 0.9575 & 0.3630 & 0.3890 & 0.3416 \\
& DNN             & 0.6239 & 0.6361 & 0.9796 & 0.4237 & 0.2840 & 0.2997  \\
\midrule
\multirow{5}{*}{Task$~$3}
& RaGSECo          & \bf{0.4464} & \bf{0.4014}& 0.8848 & \bf{0.3001} & \bf{0.2513}  & \bf{0.2600} \\
& MCFF-MTDDI      & 0.4400 & 0.3870 & 0.8701 & 0.2823 & 0.2351 & 0.2437  \\ 
& MDF-SA-DDI      & 0.4338 & 0.3873 & 0.8630 & 0.2715 & 0.2226 & 0.2329  \\
& DDIMDL          & 0.4075 & 0.3635 & 0.9512 & 0.2408 & 0.1452 & 0.1590 \\
& DeepDDI         & 0.3602 & 0.2781 & 0.9059 & 0.1586 & 0.1450 & 0.1373  \\
& DNN             & 0.4087 & 0.3776 & \bf{0.9550} & 0.1836 & 0.1093 & 0.1152  \\
\bottomrule
\end{tabular*}
\label{dataset1}
\end{table*}

\begin{figure*}[t]
\centering
\subfigure[]{\includegraphics[width=2.2in,height=2in]{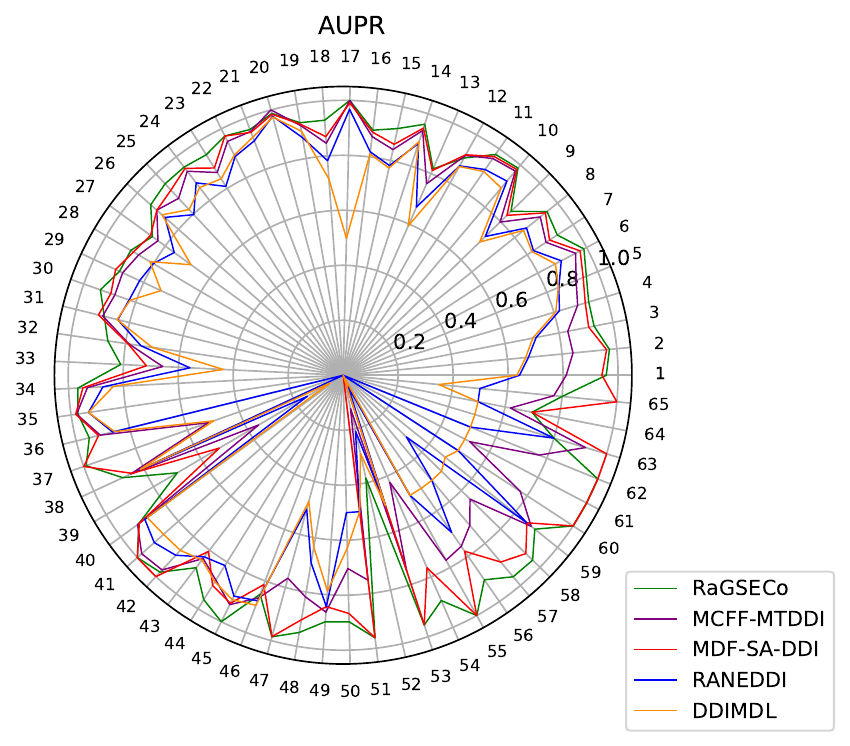}}
\subfigure[]{\includegraphics[width=2.2in,height=2in]{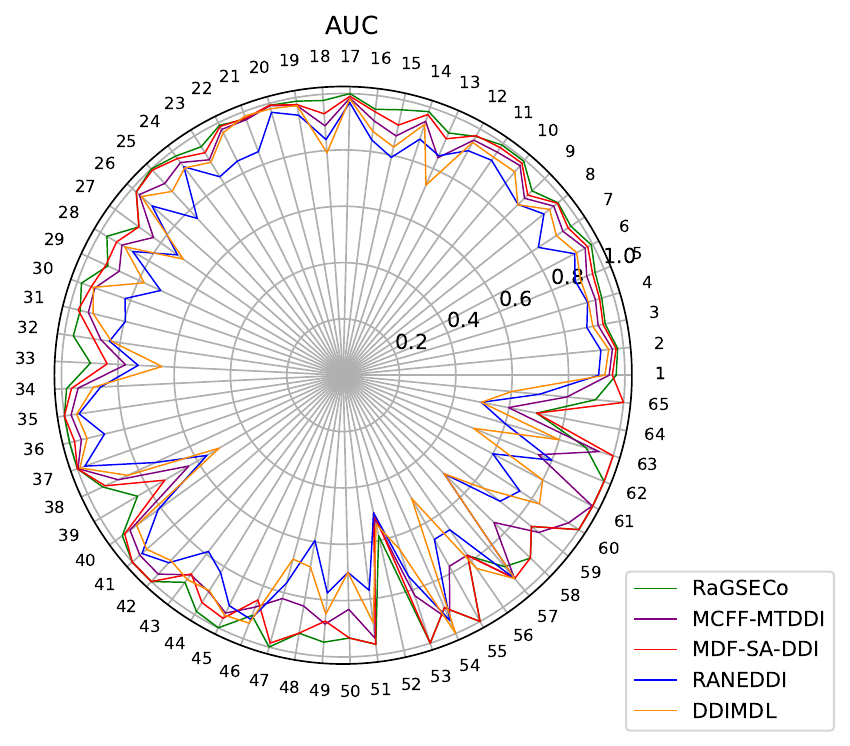}}
\caption{Performance comparison for each DDI event type of Dataset 1.}
\label{ZZ}
\end{figure*}

\begin{table*}[t]\small
\caption{Prediction performances of different methods on Dataset 2.}
\tabcolsep=0pt
\begin{tabular*}{\textwidth}{@{\extracolsep{\fill}}lccccccc@{\extracolsep{\fill}}}
\toprule%
            & Method $~$ & ACC $~$&  AUPR $~$ & AUC$~$ & Precision$~$ & Recall & F1$~$\\
\midrule
\multirow{7}{*}{Task$~$1}
& RaGSECo    & \bf{0.9344} & \bf{0.9805} & \bf{0.9995}  & 0.9021 & \bf{0.9234} & \bf{0.9113}\\
& MCFF-MTDDI & 0.9010 & 0.9532 & 0.9984  & 0.8300 & 0.9122 & 0.8631 \\ 
& MDF-SA-DDI & 0.8725 & 0.9385 & 0.9979  & 0.7518 & 0.9198 & 0.8220 \\
& RANEDDI    & 0.8611 & 0.9225 & 0.9872  & 0.8155 & 0.9084 & 0.8110 \\
& DDIMDL     & 0.8401 & 0.8824 & 0.9892  & 0.7678 & 0.8580 & 0.7800 \\
& DeepDDI    & 0.7813 & 0.8542 & 0.9810  & 0.6459 & 0.8065 & 0.6732 \\
& DNN        & 0.8281 & 0.8650 & 0.9722  & 0.7337 & 0.8239 & 0.7560 \\ 
 
\midrule
\multirow{5}{*}{Task$~$2}
& RaGSECo    & \bf{0.6570} & \bf{0.6800}& \bf{0.9862} & \bf{0.5576} & \bf{0.6087} & \bf{0.5586}\\
& MCFF-MTDDI & 0.6422 & 0.6564 & 0.9601 & 0.5407 & 0.5087 & 0.5140 \\ 
& MDF-SA-DDI & 0.6333 & 0.6486 & 0.9667 & 0.5317 & 0.4900 & 0.4785 \\
& DDIMDL     & 0.6039 & 0.6159 & 0.9737 & 0.5080 & 0.4465 & 0.4470 \\
& DeepDDI    & 0.5266 & 0.5122 & 0.9522 & 0.3030 & 0.3592 & 0.3219 \\
& DNN        & 0.5985 & 0.6035 & 0.9666 & 0.3572 & 0.2475 & 0.2661 \\
\midrule
\multirow{5}{*}{Task$~$3}
& RaGSECo    & \bf{0.4775} & \bf{0.4362} & 0.9578 & \bf{0.2957} & \bf{0.2898} &\bf{0.2723}\\
& MCFF-MTDDI & 0.4545 & 0.4136 & 0.9340 & 0.2857 & 0.2348 & 0.2427\\ 
& MDF-SA-DDI & 0.4530 & 0.4092 & 0.9358 & 0.2768 & 0.2336 & 0.2399  \\
& DDIMDL     & 0.4111 & 0.3739 & \bf{0.9514} & 0.2715 & 0.1691 & 0.1823\\
& DeepDDI    & 0.3466 & 0.2685 & 0.9165 & 0.2284 & 0.1535 & 0.1640 \\
& DNN        & 0.4062 & 0.3680 & 0.9454 & 0.1576 & 0.1245 & 0.1373  \\
\bottomrule
\end{tabular*}
\label{alldataset2}
\end{table*}

\subsection{Comparison with Other Methods}\label{Comparison}

\subsubsection{Dataset 1}\label{Dataset1}
To assess the competitiveness of our RaGSECo, we compared it with several state-of-the-art DDI prediction methods, namely MCFF-MTDDI, MDF-SA-DDI, RANEDDI, DDIMDL, DeepDDI, and DNN.
Table~\ref{dataset1} presents the metric scores achieved by these methods across the three tasks of Dataset 1.
As observed, in most cases, the comparison results demonstrate that our RaGSECo outperforms the competitors in terms of performance metrics in three tasks. 
In Task 1, although RANEDDI also considers relation-aware graph structure information to learn drug embedding and achieve outstanding prediction performance, our RaGSECo achieves better results than RANEDDI.
Specifically, the improvements of RaGSECo over RANEDDI are $2.33\%$, $1.81\%$, $3.74\%$, $2.83\%$, and $3.33\%$ in terms of ACC, AUPR, Precision, Recall, and F1, respectively. 
The reasons causing this phenomenon are manifold. On the one hand, RANEDDI ignores multiple drug-related attributes, such as targets, enzymes, and chemical substructures. This makes RANEDDI lack the ability to capture relationships beyond interactions between drugs.
On the other hand, simply concatenating embeddings of two drugs to obtain drug-pair features also limits the generalization ability of RANEDDI.
MDF-SA-DDI considers multiple attributes representing drugs and employs
multiple drug fusion methods to learn drug-pair features. 
Nevertheless, our RaGSECo still has $1.63\%$, $1.01\%$, $2.83$, and $1.72\%$ improvements over MDF-SA-DDI with respect to ACC, AUPR, Recall, and F1, respectively.
The main reason is that MDF-SA-DDI does not take advantage of specific interaction information between drugs.   

For further insight, we investigate the performances of our RaGSECo and four competitive baselines for each event.
Fig.~\ref{ZZ} displays the AUPR and AUC scores of the five prediction models for each event on Task 1.
As observed, RaGSECo produces greater AUPR and AUC scores than other methods in most event types. 
In most cases, RaGSECo can achieve a satisfactory result.
All unsatisfactory prediction results are observed in low-frequency event types, such as $\#39$, $\#52$, and $\#64$, with frequencies of only $98$, $24$, and $10$ samples, respectively.
The limited availability of training samples for these low-frequency event types may contribute to relatively poorer performance.

In Tasks 2 and 3, we compare our RaGSECo with five competitive prediction methods, i.e., MCFF-MTDDI, MDF-SA-DDI, DDIMDL, DeepDDI, and DNN. 
Since RANEDDI only focuses on the embedding of known drugs, RANEDDI can not be applied to Task 2 and Task 3.
In Task 2 and Task 3, the test DDIs include new drugs, and the lack of interaction information on the new drugs weakens the generalization ability of models.
Therefore, the prediction accuracy of models performed on Task 2 and Task 3 is lower than that on Task 1.
Nevertheless, RaGSECo outperforms other competitors in most cases. The reasons are three folds: 
1) RaGSECo enables all drugs, including new drugs, to capture effective relation-aware interaction information.
2) RaGSECo inventively combines initial features, SMILES information, and drug RaGSEs, enhancing the information diversity.
3) RaGSECo captures the underlying correlation between DPs and enables two views to supervise each other by co-contrastive learning.

\subsubsection{Dataset 2}\label{Dataset2}

Table \ref{alldataset2} presents the performance metrics of the proposed RaGSECo and other outstanding approaches on three tasks of Dataset 2.
Dataset 2 exhibits greater diversity compared to Dataset 1.
It has more drugs, more DDIs, and a wider range of DDI event types.
As observed, our RaGSECo can achieve better prediction performances than other competitors in most cases.
In Task 1, compared with MCFF-MTDDI, the proposed RaGSECo has $0.73\%$, $2.25\%$, $1.47\%$, and $2.11\%$ performance improvements in terms of ACC, Precision, Recall, and F1, respectively.
As observed in Tasks 2 and 3, our RaGSECo still acquires the best results compared with five competitive prediction models. 
The experiment results demonstrate the effectiveness and robustness of RaGSECo.

\subsection{Case Study}\label{caseexperiment}

In this section, we conduct case studies to validate the usefulness of
RaGSECo.
We adopt all DDIs and their event types of Dataset 2 to train the RaGSECo model. Then, we use the trained RaGSECo model to test the other DPs. Finally, we report the top-ranked prediction results. 
We pay attention to five events with the highest frequencies and check
up the top 10 predictions related to each event.
Then, we test the selected DPs using the DDI Checker tool\footnote{\url{https://go.drugbank.com/interax/multi_search}}. 
In the selected 50 DPs, 7 DDIs are confirmed and recorded in Table~\ref{casestudytab}.
For example, Ribavirin may decrease the excretion rate of Sofosbuvir, which could result in a higher serum level.
The metabolism of Sevoflurane can be increased when combined with Prednisolone phosphate.

\begin{table*}[t] 
\setlength\tabcolsep{3pt}\footnotesize
\caption{The drug names and event types of the confirmed DDIs.}
\begin{tabular*}{\textwidth}{@{\extracolsep{\fill}}llll@{\extracolsep{\fill}}}
\toprule
Rank  & Drug names  & DDI event type\\
\midrule
1 & Prednisolone phosphate and Sevoflurane    &  The metabolism increase  \\
2 & Dexamethasone and Amlodipine              &  The metabolism increase  \\
3 & Prednisolone phosphate and Flunitrazepam  &  The metabolism increase   \\  
4 & Gestoden and Clofazimine                  &  The metabolism decrease  \\
5 & Mometasone furoate and Cytarabine         &  The metabolism decrease \\
6 & Prednisolone phosphate and Ranolazine     &  The serum concentration increase \\
\multirow{2}{*}{$7$} & \multirow{2}{*}{Velpatasvir and Ribavirin} &The excretion rate which could result in  \\
         &   & a higher serum level decrease\\ 
\bottomrule
\label{casestudytab}
\end{tabular*}
\end{table*}

\begin{table*}[t]\scriptsize
\setlength\tabcolsep{5pt} 
\caption{The hyper-parameters of best accuracy for the proposed RaGSECo on all Tasks.}
\begin{tabular*}{\textwidth}{@{\extracolsep{\fill}}lll@{\extracolsep{\fill}}}
\toprule
Dataset & Task & Hyper-parameters \\
\midrule
\multirow{3}{*}{Dataset$~$1$~~~$}
& Task 1  & $bs$:512,  $lr$:2e-5,  $dr$:0.3,  $te$:120,  $d'$:500,  $n$:0,  $d^{FNN}$:1000,  $t_{pos}$:0.95,  $t_{neg}$:0.1, $\lambda$:5\\
& Task 2   & $bs$:512, $lr$:5e-6, $dr$:0.2, $te$:120, $d'$:500, $n$:3, $d^{FNN}$:1500, $t_{pos}$:0.95, $t_{neg}$:0.1, $\lambda$:5\\
& Task 3   & $bs$:512, $lr$:5e-6, $dr$:0.2, $te$:120, $d'$:500, $n$:3, $d^{FNN}$:1500, $t_{pos}$:0.95, $t_{neg}$:0.1, $\lambda$:5\\
\midrule
\multirow{3}{*}{Dataset$~$2$~~~$}
&Task 1    & $bs$:1024, $lr$:2e-5, $dr$:0.5, $te$:120, $d'$:500, $n$:0, $d^{FNN}$:1500, $t_{pos}$:0.95, $t_{neg}$:0.1, $\lambda$:5\\
&Task 2   & $bs$:1024, $lr$:5e-6, $dr$:0.5, $te$:120, $d'$:500, $n$:3, $d^{FNN}$:1500, $t_{pos}$:0.95, $t_{neg}$:0.1, $\lambda$:5 \\
&Task 3    & $bs$:1024, $lr$:5e-6, $dr$:0.5, $te$:120, $d'$:500, $n$:3, $d^{FNN}$:1500, $t_{pos}$:0.95, $t_{neg}$:0.1, $\lambda$:5\\
\bottomrule
\label{hyperparameters_value}
\end{tabular*}
\end{table*}

\section{Conclusion}

In multi-relational DDI prediction, relation-aware graph embedding-based methods hold considerable promise.
Nevertheless, interaction information of new drugs is unknown, which may cause these methods to suffer from severe over-fitting when predicting DDIs involving new drugs.
To address this issue, we introduce the RaGSECo approach.
The primary contribution of the RaGSECo is enabling all drugs to capture effective relation-aware interaction features and promoting the identical distribution of the training set and test set.
Additionally, our RaGSECo employs a cross-view contrastive mechanism to boost DP representation learning by utilizing the underlying correlation between DPs and supervising two views collaboratively.
Our approach offers a promising drug and DP representation learning solution, thereby enhancing DDI prediction performance.


\section*{Declaration of Competing Interest}
\label{Declaration}
The authors declare that they have no known competing financial interests or personal relationships that could have
appeared to influence the work reported in this paper.







\end{document}